\documentclass[lettersize,journal]{IEEEtran}
\usepackage{amsmath,amsfonts}
\usepackage{algorithmic}
\usepackage{algorithm}
\usepackage{array}
\usepackage[caption=false,font=normalsize,labelfont=sf,textfont=sf]{subfig}
\usepackage{textcomp}
\usepackage{stfloats}
\usepackage{url}
\usepackage{verbatim}
\usepackage{graphicx}
\usepackage{cite}

\usepackage{booktabs}
\usepackage{rotating,tabularx}
\usepackage{xcolor}

\begin{document}

\title{A Survey on Drowsiness Detection -- Modern Applications and Methods}


\author{{Biying Fu, Fadi Boutros, Chin-Teng Lin and Naser Damer}}



\maketitle

\begin{abstract}
Drowsiness detection holds paramount importance in ensuring safety in workplaces or behind the wheel, enhancing productivity, and healthcare across diverse domains. Therefore accurate and real-time drowsiness detection plays a critical role in preventing accidents, enhancing safety, and ultimately saving lives across various sectors and scenarios. This comprehensive review explores the significance of drowsiness detection in various areas of application, transcending the conventional focus solely on driver drowsiness detection. We delve into the current methodologies, challenges, and technological advancements in drowsiness detection schemes, considering diverse contexts such as public transportation, healthcare, workplace safety, and beyond. By examining the multifaceted implications of drowsiness, this work contributes to a holistic understanding of its impact and the crucial role of accurate and real-time detection techniques in enhancing safety and performance. We identified weaknesses in current algorithms and limitations in existing research such as accurate and real-time detection, stable data transmission, and building bias-free systems. Our survey frames existing works and leads to practical recommendations like mitigating the bias issue by using synthetic data, overcoming the hardware limitations with model compression, and leveraging fusion to boost model performance. This is a pioneering work to survey the topic of drowsiness detection in such an entirely and not only focusing on one single aspect. We consider the topic of drowsiness detection as a dynamic and evolving field, presenting numerous opportunities for further exploration.
\end{abstract}


\section{Introduction}
\begin{figure}
    \centering
    \includegraphics[width=0.98\linewidth]{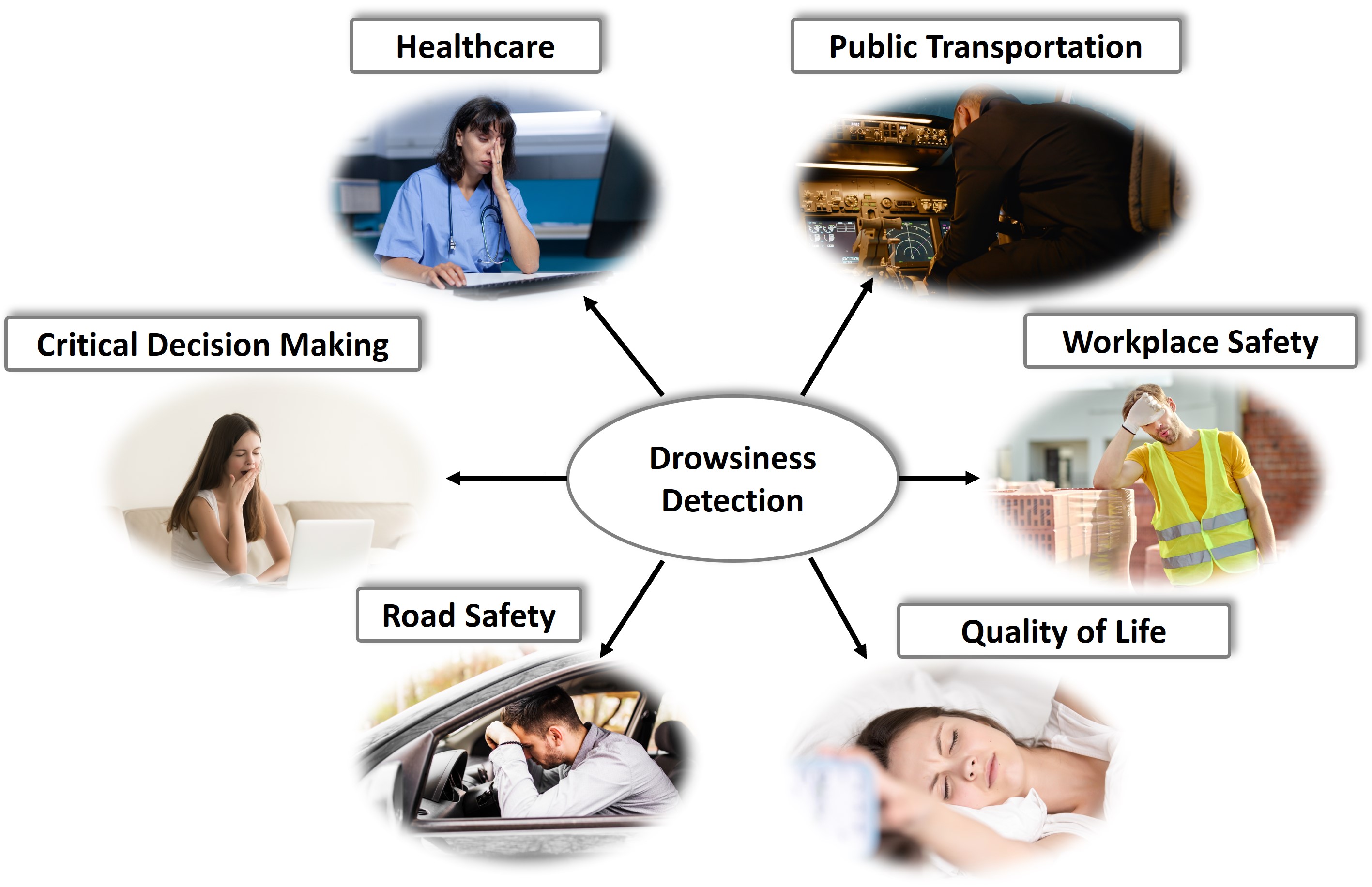}
    \caption{Depicts areas of applications in urgent need of accurate and real-time drowsiness detection task.}
\label{fig:drowsiness_applications}
\end{figure}

Drowsiness detection aims at detecting the early symptoms of individual drowsiness using physiological (EEG \cite{DBLP:journals/ijns/QinNZLJZ23}, ECG \cite{Lee_Lee_Shin_2019}) and visible behavioural (eye blinking \cite{deng2019real} and eye closure \cite{garcia2012vision}) indications. The accurate and real-time identification of drowsiness detection is crucial for multiple reasons. One of the most prominent use-cases is driver and road safety; drowsy driving \cite{Sahayadhas_Sundaraj_Murugappan_2012} is a dominant cause of accidents, injuries, and fatalities on the road. The National Highway Traffic Safety Administration estimated that up to 20\% of the annual traffic deaths were attributed to driver drowsiness in 2016 \cite{national2016nhtsa}. An early alert can help prevent accidents caused by impaired reaction times of a drowsy driver. Under the setting of workplace safety, such as employees working in jobs like e.g., operating heavy machinery \cite{LIU2021103901}, industrial equipment, or in medical establishment \cite{Geiger-Brown_Rogers_Trinkoff_Kane_Bausell_Scharf_2012,mantzanas2022subjective, Juan-García_Plaza-Carmona_Fernández-Martínez_2021}, drowsiness can lead to accidents that jeopardize worker's safety or patient safety. In addition, drowsiness could negatively impact the cognitive function and productivity of shift workers with long working hours. Even in healthcare, monitoring the drowsiness of patients, especially those with sleep disorders or undergoing medical treatments, ensures their well-being and helps healthcare providers make informed decisions about accurate treatment plans \cite{woodard20220064,kamran2019drowsiness,papadelis2007monitoring,boyle2008driver,Sharma_Sharma_2016}. Finally, drowsiness detection is crucial in sectors such as aviation \cite{wang2019spectral,Jeong_Yu_Lee_Lee_2019} and public transportation \cite{Zhang_2017,chen2022driver,zhou2023driver}, where a drowsy operator can compromise passenger safety. 
The attentiveness of the driver significantly impacts railway safety \cite{chen2022driver}. Incidents involving high-speed trains can result in severe consequences, as evidenced by the most catastrophic railway accident in Chinese history occurring on 23 July 2011, resulting in 40 fatalities and at least 192 injuries \cite{fan2015applying}. Therefore, finding ways to unobtrusively detect driver drowsiness operating these high-speed trains is important. Figure \ref{fig:drowsiness_applications} summarizes the application areas that can benefit greatly from accurate and real-time drowsiness detection. 

Drowsiness detection is thus vital for safety, preventing accidents caused by impaired attention and reaction times. It improves health by identifying sleep disorders and enhances productivity in various settings. After motivating the importance of drowsiness detection, our work preliminary focused on showing modern applications and methods on drowsiness detection. We are the first work concluding multiple aspects and areas of drowsiness detection not only focusing especially on driver drowsiness detection. Current surveys on drowsiness detection mainly focused on presenting either techniques \cite{Ramzan_Khan_Awan_Ismail_Ilyas_Mahmood_2018, More_2022, Reddy_Kim_Yun_Seo_Jang_2017,othmani2023eeg} or systems \cite{s121216937,Sahayadhas_Sundaraj_Murugappan_2012,el2024machine} on driver drowsiness detection. Less focus was put on the general use-cases of drowsiness detection in such a broad application area and its faced challenges. 

\begin{figure*}
    \centering
    \includegraphics[width=0.98\textwidth]{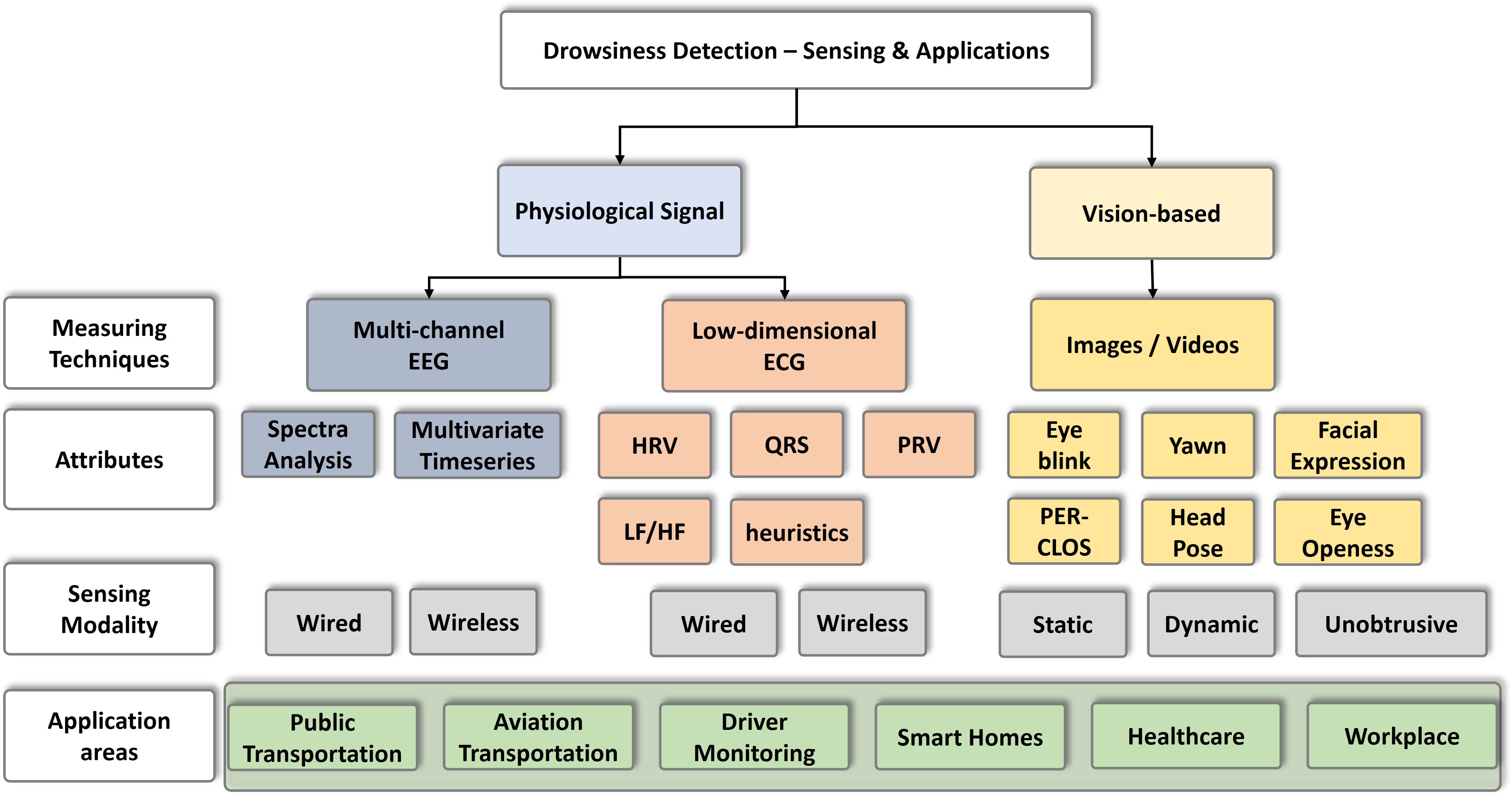}
    \caption{Overall structure of the surveyed advancements in drowsiness detection. This primary framework illustrates the three measuring techniques utilized in drowsiness detection, along with their respective attributes and methodologies. From the considered topology, we first distinguish between drowsiness detection based on physiological signals and monitoring through vision. Based on these classification, we assign different attributes to individual measuring techniques. The sensing modalities for EEG and ECG are not mutually exclusive and can be divided under wired and wireless application. While under the vision-approach, we now consider static investigation for single frame and dynamic investigation for multiple frames and the unobtrusiveness for the general setup. Abbreviations like HRV stands for heartrate variability, PRV for pulserate variability, QRS for morphological components of a heartbeat, LF/HF for certain frequency bands, and finally PERCLOS for percentage of eye closure. The sensing modality demonstrates the potential realization of applications. The bottom row outlines the typical areas of application for drowsiness detection across all measuring techniques.}
\label{fig:drowsiness_category}
\end{figure*}

\newpage
Figure \ref{fig:drowsiness_category} depicts a primary framework of this survey, illustrating the three measuring techniques and their respective attributes used for detection, sensing modalities, and typical application areas across all measuring techniques. and provides a clear view of the structure of our work. Based on our proposed topology, we differentiate between drowsiness detection relying on precise physiological signals and monitoring through vision-based approaches. Biological processes manifest themselves in brain and heart activities, that can lead to distinct changes in specific attributes induced by drowsiness. Consequently, we can assign different attributes to individual measuring techniques based on our classification. The sensing modality for physiological signals typically involves electrode-based methods. However, there is a shift towards wearable and wireless communication technologies in this regard to enhance user acceptance. Monitoring through vision is in general more unobtrusive and user friendlier, focusing on relevant attributes such as facial features, eye closures, and head poses, among others. We view single frame as static and consecutive frames as dynamic sensing modality in this context. 

The structure of our work is outline in the following. In Section \ref{sec:area_of_applications}, we discuss recent works within each of the application areas. In Section \ref{sec:technology} we first introduce three general techniques typically used for drowsiness detection under two main categories, i.e. physiological signal- and vision-based approaches. In Section \ref{sec:modern_applications}, we extend the techniques with relevant works. In Section \ref{sec:database} we provide a list of public benchmarks used to evaluate or design algorithms for drowsiness detection both in form of time series and image- or video-sequence-based data. In Section \ref{sec:evaluation}, we exclusively explained the performance and evaluation metrics used for assessing the algorithm's effectiveness in state-of-the-art (SOTA) works. We then discuss and reveal the existing limitations within this research area categorized under physiological and vision-based approaches and present potential solutions and actionable suggestions in Section \ref{sec:discussion}. In Section \ref{sec:future_direction}, we summarize the current research gaps and provide potential future research directions. Finally, we conclude our work by summarizing the major findings in Section \ref{sec:conclusion}.

\section{Areas of Applications}
\label{sec:area_of_applications}

In this section, we first start with relevant works for drowsiness detection grouped by applications. Beyond the scope of driver's drowsiness detection, drowsiness detection can take place in several other aspects of our daily life, as it has far-reaching effects ranging from affecting our quality at work, influencing our healthcare, or clouding our ability at working on cognitive tasks.

\paragraph{Drowsiness detection in workplace and secured areas}
Drowsiness and sleepiness in the workplace are two of the major risks of modern society \cite{CALDWELL2019272,b584016b-ce16-3669-a4a1-006ccfeb28ff}. Well-rested and alert employees are fundamental for better productivity and creativity, while excessive fatigue not only reduces efficiency but also can pose a risk at workplaces, thus fatigue management in the workplace \cite{sadeghniiat2015fatigue} aims to enhance worker health and well-being both on and off the workspace. 

One example of a mitigating safety issue caused by drowsiness at the workplace is proposed in \cite{LIU2021103901}. The work by Liu et al. \cite{LIU2021103901} focused on improving the performance of drowsiness detection for crane operators by using hybrid deep neural networks connecting both spatial and temporal features from videos. The primary contributions of this study involved extending drowsiness detection beyond vehicle drivers to crane operators, introducing relevant facial features as indicators for detection. In addition to their customized dataset on simulated crane operation scenarios, the research provided recommendations for gathering extensive and publicly accessible realistic drowsiness datasets tailored to crane operators.

Sectors engaged in safety-critical endeavors, including the oil and gas industry, exhibit a vested interest in monitoring biological markers to avert human errors and enhance process safety, thereby enhancing their readiness for emergency situations \cite{astaras2008biomedical}. Within this context, the reliability of human performance assumes a pivotal role in preventing potential catastrophic incidents stemming from human factors, such as worker fatigue. Ramos et al. \cite{Ramos_Maior_Moura_Lins_2022} introduced an ensemble approach for fully automated drowsiness detection utilizing Electroencephalogram (EEG) signals. 

Even in office environments across various industries, persistent drowsiness poses a significant challenge, resulting in occupational fatigue due to excessive work demands, particularly when precise operational output is essential. Presently, companies are increasingly dedicating resources to monitor their workplaces, aiming to uphold employees' optimal working states and sustain a desirable level of productivity \cite{CALDWELL2019272}. Natnithikarat et al. \cite{natnithikarat2019drowsiness} proposed one approach to detect the drowsiness state of an office worker that combines Biometric characteristics like keyboard keystrokes and mouse movement along with eye tracking during office-related tasks. The objective was to identify and assess drowsiness based on the self-reported Karolinska sleepiness scale (KSS) \cite{kaida2006validation} questionnaire. The study identified a correlation between the anticipated level of drowsiness inferred from the biometric data and the estimated KSS score provided by the users. These findings suggest the viability of using the proposed method to effectively detect the level of drowsiness among office workers.

\paragraph{Drowsiness detection in healthcare sectors}

Modern industries, hospitals, and many other essential sectors require shift work to maintain productivity and profit\cite{ummul2012shift, DBLP:journals/hf/ThompsonSB17}. However shift work with its induced drowsiness and fatigue in shift workers has proved to be the source of human error and can lead to a number of accidents, catastrophes, and health-related problems \cite{DBLP:journals/hf/ThompsonSB17,aakerstedt1988sleepiness}.

In healthcare sectors, Geiger-Brown et al. \cite{Geiger-Brown_Rogers_Trinkoff_Kane_Bausell_Scharf_2012} revealed in their study that nurses frequently express fatigue and dissatisfaction with the quality of sleep when engaged in 12-hour shifts. This study examines the sleep patterns, levels of sleepiness, fatigue, and neuro-behavioral performance across three consecutive 12-hour shifts (both day and night) for hospital nurses. The findings indicate that nurses accumulate a substantial sleep deficit during successive 12-hour shifts, leading to increased fatigue, sleepiness, and decreased attention. Other studies investigating the subjective sleep quality and daytime sleepiness among nursing staff can be found in \cite{mantzanas2022subjective, Juan-García_Plaza-Carmona_Fernández-Martínez_2021}. As a result, it becomes imperative to promptly and accurately identify fatigue in real-time to mitigate potential instances of malpractice stemming from fatigue-related issues in the medical domain.

Chervin et al. \cite{chervin2000sleepiness} conducted an investigation in 2000, focusing on the relation of sleepiness, fatigue, tiredness, and lack of energy in individuals to obstructive sleep apnea. The study encompassed a comprehensive analysis involving 190 participants, including 117 males (M) and 73 females (F), within a university-affiliated sleep laboratory. Data were derived from both sleep studies and questionnaires. The study's findings suggested that complaints regarding fatigue, tiredness, and reduced energy levels could hold comparable significance to reports of sleepiness among obstructive sleep apnea patients. Notably, female patients seemed to express such concerns more frequently than their male counterparts. 

\paragraph{Drowsiness detection at public transportation and aviation}

In the public transportation sector, accurate and real-time drivers' drowsiness detection can save human lives \cite{zhou2023driver}. Wu et al. \cite{wu2021fatigue} proposed a non-parametric solution for detecting the cognitive state of pilots by utilizing a 64-channel brain EEG signal. They developed 2D brain maps from these spatially distributed 3D multichannel EEG signals and extracted useful feature maps for developing algorithms to detect fatigue in pilots. Another work by Wang et al. \cite{wang2019spectral} showcased drowsiness detection within the aviation industry too. They utilized EEG signals obtained from a standard aviation headset to identify the drowsiness levels of pilots. The researchers integrated the Seeing Machines driver monitoring system 
\footnote{Seeing Machines driver monitoring system: https://seeingmachines.com/} with electrooculogram (EOG) data to localize microsleep events and investigated unique characteristics in EEG spectral patterns during these events. Simultaneous recordings of EEG, EOG, and facial behavior data were taken from 16 pilots during simulated flights. Their study demonstrated useful features from the EEG signals and confirmed the effectiveness of drowsiness detection by embedding EEG electrodes within the commonly used aviation headset.

Zhang et al. \cite{Zhang_2017} targeted drivers operating high-speed trains. They proposed a driver's drowsiness detection system for high-speed train safety based on monitoring train driver's vigilance using a wireless wearable EEG. The proposed system includes three stages ranging from wireless data collection and driver vigilance detection to pushing early alert messages to drivers. An 8-channel wireless wearable brain-computer interface is used to unobtrusively collect the locomotive driver's brain EEG signal, while the driver is simultaneously operating a high-speed train.

Multi-modal fusion of multiple physiological signals and leveraging deep learning techniques for driver's drowsiness detection for high-speed rail operators can be found in \cite{chen2022driver}. One research specifically deals with drowsiness detection from facial clues with occluded face images of railway drivers \cite{zhou2023driver}. This study was conducted in the post-covid phase to work with the autonomous-rail rapid transit system in China railway. This research was built upon facial thermal imaging and also included environmental information for detection.

\paragraph{Drowsiness Detection in smart home context}
Beyond workspace applications, modern homes with integrated smart technologies also intend to enhance the user's living experience in domestic areas. The smart home application with the smart mirror is used as an example to assess residents' well-being over time to improve their lifestyle through user-centered guidance. Facial clues indicating user's emotional states like stress, fatigue and anxiety are targeted in \cite{henriquez2017mirror}. 

Elderly fall due to drowsiness was studied in \cite{kumar2021elderly}. Based on facial landmarks and eye openness, Kumar et al. further analyzed sleep patterns in order to help to predict the physical condition of the elderly and to avoid emergency situations such as falls. The main contribution of this work is to predict the health condition of the elderly by leveraging machine learning models and their results were verified on real-world scenarios while maintaining good accuracy (Acc).

Another proof of concept work focusing on real-time drowsiness detection for elderly care is in \cite{9371810}. The study is based on video feeds to extract visible facial clues, such as yawning, eyelid and head movement over time which are related to drowsiness detection. Classification simply based on eyelid and mouth status already achieved an accuracy between 94.3\%-97.2\%.

In this survey, we focus on showing modern application areas and methods in terms of drowsiness detection not only limited to driver's drowsiness detection but extended to cover much broader possible scenarios and detection under more general and diverse purposes.

\section{Measuring Technology}
\label{sec:technology}

In this section, we present the three most popular measuring techniques for drowsiness detection. These measuring techniques can be categorized under physiological sensing (which uses either EEG or ECG) and vision-based sensing. Under physiological sensing, researchers aim to capture numerous biological signals of each individual, such as heart rate variability, muscle movement, and brain wave activities. From these biological signals, unique biological markers are extracted to detect drowsiness. Under vision-based capture, visual feeds are leveraged to derive facial expressions and eye blinking status, among other clues, to give an indication of drowsiness. 

\subsection{How does Electroencephalogram work for drowsiness detection?}
Electroencephalogram (EEG) \cite{teplan2002fundamentals} measures the electrical activity of the brain. It involves the recording of the brain's electrical signals using electrodes placed on the scalp. EEG is a non-invasive and painless procedure used to study brain activity, diagnose certain brain disorders, and monitor brain health during medical treatments.

EEG-based drowsiness detection systems are just one component of an overall driver safety system. During a driving simulation or driving a vehicle in a controlled environment, the EEG system continuously records brain activity throughout the driving session. From the raw multi-channel EEG data, relevant features are extracted that correlate with drowsiness. These features often include changes in different brainwave frequencies \cite{ jiang2019robust}, asymmetry in frequency bands \cite{da2016automated}, or power spectral density \cite{geering1993period, DBLP:conf/siu/OnayK19}. Similarly, these features are used to train a machine learning model detecting the real-time state of the driver as in either awake or drowsy state. Mardi et al. \cite{mardi2011eeg} demonstrated that the brain exhibits its lowest levels of activity and complexity when a driver experiences drowsiness. Consequently, individuals in such a state lose their concentration and control, thereby hindering their ability to respond promptly to stimuli.

EEG is a valuable tool in neuroscience and clinical settings, providing insights into brain function and helping diagnose and manage various neurological conditions. Its non-invasiveness and ability to capture real-time brain activity make it a widely used method for studying brain health and understanding brain-related disorders. EEG also serves as the 'gold standard' and is extensively applied to indicate the transition between wakefulness and sleepiness \cite{jiao2020driver}. Spontaneous alpha activity detected in EEG signals can indicate different underlying physiological process. Spontaneous alpha activity refers to rhythmic electrical oscillations in the brain that occur predominantly in the alpha frequency range, typically between 8 and 13 Hz \cite{cantero2002human}. According to Cantero et al. \cite{cantero2002human}, alpha activity in this range can be related to processes like wakefulness, drowsiness period, and REM sleep phase. Ogilvie et al. \cite{ogilvie2001process} investigated the process of falling sleep in relation to the alpha activities being recorded and associated the alpha wave disappearance to be related to stage I sleep. Kleitman and his students \cite{blake1939factors} also examined the correlation between muscle relaxation and EEG activity, observing a drop of a hand-held spool between 0.5 and 25 seconds after the alpha wave had vanished.


\subsection{How does Electrocardiogram work for drowsiness detection?}

The most effective outcomes in driver drowsiness detection have been attained through these electrode based instruments including EEG and EOG measurements to date\cite{heinze2017drowsiness}, thus establishing them as the prevailing standard in this research domain. Nevertheless, implementing driver monitoring using EEG measurement proves impractical. In contrast, Electrocardiogram (ECG) recording offers a more feasible alternative, given its ease of capture, significantly larger magnitude, and lower susceptibility to noise interference.

Sleep is a complex state marked by important changes in the autonomic modulation of the cardiovascular activity as investigated by Viola et al. \cite{viola2011short}. Heart rate variability (HRV) undergoes substantial alterations across different sleep stages, reflecting a prevailing parasympathetic influence on the heart during non-rapid eye movement (NREM) sleep, while displaying heightened sympathetic activity during rapid eye movement (REM) sleep. In addition, respiration also undergoes notable changes, deepening and becoming more regular during deep sleep and shallower and more frequent during REM sleep. These effects were thoroughly investigated by Cabiddu et al. in \cite{cabiddu2012modulation} and thus making the ECG measurement a well-suited technique for investigating sleep analysis and drowsiness detection.

ECG is a medical instrument to record the electrical activity of the heart over a period of time. The ECG provides valuable information about the heart's rhythm, rate, and overall electrical activity \cite{alghatrif2012brief}. To perform an ECG, a set of electrodes is placed on the patient's skin at specific locations. The electrodes are connected to an electrocardiograph which detects and amplifies the electric signals generated by the heart. The recorded electrical impulses are time series representing the movement of the heart's chambers during each heartbeat. A cardiologist or a trained technician can analyze the ECG recording to identify abnormalities or irregularities that can indicate various heart conditions or problems. ECG signals can exhibit various abnormalities indicative of heart related deceases, including arrhythmia \cite{hammad2018detection}, ST-segment abnormalities \cite{brady2001cause} as indicators for hear attack, T-wave abnormalities \cite{kumar2007clinical} and  QT interval prolongation \cite{trinkley2013qt} indicative of electrolyte imbalances, QRS complex abnormalities \cite{strauss2009qrs} indicative for bundle branch blocks, and other artifact or noise \cite{kher2019signal} among others.

The ECG is a non-invasive and painless measuring procedure, making it a widely used tool for diagnosing heart-related issues. In sleep studies, ECG can be used to record heart rate variations providing hints for sleep disorders \cite{stein2012heart,penzel2016modulations} or sleep apnea \cite{khandoker2008support,DBLP:conf/embc/Papini0MGOBV18, Sharma_Sharma_2016} detection. Nowadays, simplified versions of ECGs exist, such as portable ECG devices \cite{DBLP:conf/ict/DharmaSMEC23} based on Bluetooth transmission or wearable sensors \cite{ramasamy2018wearable, DBLP:journals/sensors/LiAZZZZZSMMLG21} for ECG measurement, making the application of ECG in everyday life possible. There are several studies linking ECG-based features to drowsiness detection as in \cite{kamran2019drowsiness,papadelis2007monitoring,boyle2008driver}. Most of these features are based on detecting heart rate variability and are commonly derived from the frequency and spectrum domains.

\subsection{How does a vision-based drowsiness detection system work?}

Both previous methods, which rely on biosignals, are well-suited for laboratory conditions but are not very practical in real-world driving scenarios. This is because, while driving on the road, both the movement and the dynamic environment significantly impact performance and pose challenges to extracting features from captured biosignals. Both measures would require individuals sitting still and wearing cables or head-mounted devices. Thus, we motivate for vision-based approaches overcoming the restrictions of wired installation or head-mounted devices, which might effect usability in some use-cases. Vision-based drowsiness detection \cite{garcia2012vision} refers to a technique that uses visual information, such as facial expressions and eye movements, to identify signs of drowsiness in individuals \cite{ahmed2021intelligent}. This method is commonly employed in various domains, including driver fatigue monitoring \cite{deng2019real}, operator alertness assessment \cite{LIU2021103901}, and other safety scenarios \cite{wang2019spectral,Jeong_Yu_Lee_Lee_2019,Zhang_2017} where detecting drowsiness is crucial for safety and performance. Research linking facial muscle movement to different levels of muscle fatigue can be found in \cite{uchida2018identification}.

The pipeline of vision-based drowsiness detection often includes the following steps. During the image/video acquisition stage, a camera mostly installed on dashboards captures real-time visual data from the occupant, such as facial images or eye movement. Relevant facial features are extracted from the acquired images or video frames. Typical features involve facial landmarks, eye closures, head movements, and changes in gaze direction. These features are used to train a machine learning model often leading to a binary classification model to determine the binary states of awake or drowsiness \cite{ahmed2021intelligent, DBLP:journals/access/KhanNKHR23}.

Vision-based drowsiness detection systems offer real-time monitoring capabilities and can be integrated into various applications, such as in-vehicle driver assistance systems \cite{khandoker2008support} or workplace safety monitoring \cite{LIU2021103901}. They play a crucial role in enhancing safety and reducing the risk of accidents caused by drowsy or fatigued individuals.

\section{Modern Applications and Methods}
\label{sec:modern_applications}

In this section, we provide a comprehensive survey of recent relevant research, which is divided into three main methods for detecting drowsiness using either physiological signals or visual sampling. We distinguish between these methods based on the variables measured. ECG signals use low-dimensional heartbeat signals, EEG signals use multichannel brain activity signals, and visual feeds provide image data. Each method is summarized with a table containing the investigated research works, which is later discussed. Thereby, we consider various aspects, such as year of publication, area of use, specific algorithms developed, evaluation database and its properties, performance, and conclude with special remarks.

\subsection{EEG-based Drowsiness Detection}

EEG signals are often used to detect the mental stress of patients \cite{katmah2021review} but are also one of the physiological signals used to derive the drowsiness state \cite{stancin2021review}. For detecting the EEG signals, electrodes are detached from the skin directly thus allowing clear signal acquisition. Earlier works as in \cite{Zhang_2017,wang2019spectral,Ramos_Maior_Moura_Lins_2022} applied traditional machine learning approaches with handcrafted features extracted from these physiological signals. The development in the last few years moved to more advanced approaches based on Deep Q-Learning \cite{DBLP:journals/tetci/MingWWSL21} or deep learning in general \cite{Jeong_Yu_Lee_Lee_2019,DBLP:journals/ijns/QinNZLJZ23,Chaabene_Bouaziz_Boudaya_Hökelmann_Ammar_Chaari_2021}.

Traditional machine learning uses handcrafted feature extracted from the EEG power spectrum density to build efficient models for drowsiness detection. Zhang et al. \cite{Zhang_2017} applied a support vector machine (SVM) on Fast Fourier Transformation (FFT) features as a binary classifier. Wang et al. \cite{wang2019spectral} investigated unique characteristics from EEG spectral patterns during micro-sleep events. Ramo et al. \cite{Ramos_Maior_Moura_Lins_2022} leveraged data from five diverse EEG signal channels and employed ensemble learning techniques such as bagging to construct a robust and more precise drowsiness detection system. The efficacy of the system was validated using the DROZY database \cite{massoz2016ulg}. 

Spatio-temporal convolution served as the cornerstone for successive deep learning-based approaches to derive both the sequential and spatial characteristic features from this physiological signal. Jeong et al. \cite{Jeong_Yu_Lee_Lee_2019} extended the binary classification task of the drowsiness state to a more fine-grained classification of five drowsiness levels from EEG signals. They stated to be the first work providing such a detailed classification of drowsiness levels using only EEG signals. They acquired EEG data from ten pilots in a simulated night flight environment. They proposed a deep spatio-temporal convolutional bidirectional long short-term memory network (DSTCLN) model. The classification performance is evaluated using the Karolinska sleepiness scale \cite{kaida2006validation} for two mental states and five drowsiness levels. Results demonstrated the feasibility of their proposed fine-grained drowsiness classification.

Similarly, Cui et al. \cite{cui2021subject} improved the subject-independent drowsiness recognition from single-channel EEG with an interpretable CNN-LSTM model. In this work, authors put more effort on the explainability of the proposed deep learning models by revealing the network's decision with respect to the input data. Results showed a model average accuracy of 72.97\% on 11 subjects for leave-one-out subject-independent detection on a public dataset \footnote{Project page with access to data: https://figshare.com/articles/dataset/Multi-channel\_EEG\_recordings\_during\_a\_sustained-attention\_driving\_task/6427334\label{fn:eeg}}. They stated that their proposed method surpasses conventional baseline methods and other SOTA deep learning methods till publication. Similar network architecture is leveraged by Lee et al. \cite{lee2023lstm} in a more recent work. This work investigated the optimal length of input time series for more accurate detection of drowsiness at multiple levels (awake, sleep, and drowsy), while few studies have seriously considered this feature before.

Paulo et al. \cite{Paulo_Pires_Nunes_2020} investigated two approaches for drowsiness detection using EEG signals during a sustained-attention driving task. The study focused on pre-event time windows and addressed the challenge of cross-subject zero calibration. EEG signals are known for their low signal-to-noise ratio and individual differences between subjects, thus requiring individual calibration cycles. To tackle this issue, the researchers employed spatio-temporal image encoding representations in the form of recurrence plots for classification using deep CNN. The results obtained from a public dataset of 27 subjects showed the effectiveness of their cross-subject zero calibration approach, highlighting its success in drowsiness detection. Similarly, Jiang et al. \cite{DBLP:journals/tits/JiangZ0WL21} targeted the same issue by only requiring a few subject-specific calibrations to adjust for a new subject. They provided an online and multiview setup to enforce the consistencies across different views in both source and target domains, and thus, making the system in general more robust. In addition, online training makes the proposed application more suitable for practical requirements. Recent follow-up work focusing on cross-subject investigations was presented by Cui et al. \cite{cui2022eeg}. In this work, the focus further lies in the interpretability of drowsiness detection schemes and automatic feature selections from EEG signals.

A reinforcement learning-based method for the task of drowsiness detection is introduced in \cite{DBLP:journals/tetci/MingWWSL21}. Ming et al. \cite{DBLP:journals/tetci/MingWWSL21} leveraged deep Q-learning to analyze EEG dataset collected during simulated driving to estimate driver drowsiness state. The main research is to relate certain characteristics of the EEG data to better derive the response time in order to indirectly estimate the driver's drowsiness state. Their results showed superior performance compared to supervised learning and is promising for real applications.  

The most current research by Zhuang et al. \cite{zhuang2023connectivity} leveraged Graph Neural Networks (GNNs) for EEG-based driver drowsiness detection in real-time. Their results surpassed the accuracy of other CNNs and graph generation methods based on drowsiness detection schemes. They proposed a GNN-based network with a self-attention mechanism that can focus on developing task-relevant connectivity networks via end-to-end learning. In addition, the authors leveraged a squeeze-and-excitation (SE) block to select the most relevant features and feature bands for drivers' drowsiness detection. This block is shown both to improve the classification accuracy and the model's interpretability. 

From the data augmentation point of view, Chaabene et al. \cite{Chaabene_Bouaziz_Boudaya_Hökelmann_Ammar_Chaari_2021} introduced an EEG-based drowsiness detection system based on deep learning networks. The system follows a two-stage framework, encompassing (i) data acquisition and (ii) model analysis. For data collection, the authors employed a wearable Emotiv EPOC + headset \cite{duvinage2013performance}, recording EEG signals from 14 channels along with signal annotations. Data augmentation techniques were implemented to prevent the proposed model from overfitting. The chosen deep learning architecture in this study used a CNN network. A self-collected dataset containing 42 records of six men and eight women aged between 14 and 64 with normal mental health are used for evaluation. The outcomes exhibited a noteworthy accuracy of 90.42\% in binary classification for distinguishing drowsy and awake states. Compared to alternative research, the proposed approach demonstrated its efficacy and efficiency.

From the distributed system point of view, Qin et al. \cite{DBLP:journals/ijns/QinNZLJZ23} proposed a driver's drowsiness state detection system using EEG signals. They increased the accuracy of their system by using Federated Learning (FL) and CNN. FL is used to accumulate knowledge from the data of different clients under privacy protecting mechanism and CNN is used to identify and explain the driver's drowsiness state. However, they evaluated their method only on a relatively small amount of private database consisting of 11 subjects. 

Another more difficult and pressing issue is the detection of microsleep events (MSE). MSE \cite{golz2007feature} refers to abrupt and non-anticipated lapses of attention experienced by individuals, typically resulting from drowsiness and monotony. MSE can serve as objective indicators of excessive daytime sleepiness and can be characterized by a non-anticipated brief period of sleep lasting between 2 and 30 seconds, occurring amidst ongoing wakefulness as investigated by Carskadon et al. \cite{carskadon1993encyclopedia} in the Encyclopedia of sleep and dreaming. Microsleep accounts for an annual loss of nearly 150 million dollars due to diminished daily work performance and vehicular accidents \cite{american2016economic}. Assessing an individual's level of sleepiness and detecting the onset of microsleep is thus crucial for tasks demanding sustained focus \cite{dehart1993twenty}, such as driving or operating machinery during nighttime hours, where falling asleep poses high risks. In recent years, this subject has received widespread attention from governmental bodies, the public, and the research community alike \cite{golz2001application}.

Such subtle events like microsleep episodes are very hard to recognize. The group of Golz et al. \cite{golz2007feature} worked extensively on detecting microsleep episodes from EEG and EOG data. To achieve detection, signals coming from heterogeneous sources are processed, such as the brain electric activity captured by EEG data, variation in the pupil size, and eye and eyelid movements captured by EOG data. By combining the spectral and the state space, both linear and non-linear features are considered. The binary decision networks between MSE and non-MSE are based on a support vector machines (SMV) and a learning vector quantization (LVQ) scheme. However, pupil adaptation through light stimuli could affect the accuracy of detection.

Pham et al. \cite{pham2020wake} proposed a more flexible and mobile application by introducing WAKE, which is a behind-the-ear wearable device for microsleep detection. They utilized bone-conduction headphones to gather biosignals including brain activity, eye movements, facial muscle contractions, and galvanic responses from the area behind the user's ears. Their findings demonstrated that WAKE effectively suppressed motion and environmental noise in real-time by 9.74-19.74 dB during activities such as walking, driving, or being in various environments, ensuring reliable capture of the biosignals. A preliminary training conducted on 19 sleep-deprived and narcoleptic subjects demonstrated an average precision and recall of 76\% and 85\% on an unseen subject with leave one subject out cross validation technique.

A way to detect microsleep with deep learning architectures even with less training data is proposed by Chougule et al. \cite{9997232}. Their model uses the attention-based mechanism, which combines the advantages of the wavelet transform with the Short Time Fourier Transform (STFT) Spectogram. By separating "time-dependent" and "time-independent" parts, the deep learning model is more robust to capture both the sequence features and simultaneously learn the relationships between epochs. Only a single electrode EEG signal was used to achieve greater social acceptability. The training and evaluation are performed on the public Maintenance of Wakefulness Test (MWT) dataset \footnote{MWT dataset: https://sites.google.com/view/utarldd/home\label{fn:mwt}}.

Table \ref{tab:EEG} summarizes the investigated works using EEG signals for drowsiness detection in various application scenarios. Sensing modality covers both wired and wireless wearable devices targeting non-intrusive applications. References to the public databases are provided in the footnote. From Table \ref{tab:EEG}, we notice that all recent works starting from 2021 utilized deep learning-based approaches to mitigate the investigation of handcrafted features. We further observed that the most predominant evaluation dataset for physiological signal-based drowsiness detection is the DROZY dataset.

\begin{table*}[]
\caption{summarizes recent works performing drowsiness detection based on EEG signals.}
\label{tab:EEG}
    \centering    
    \resizebox{\textwidth}{!}{%
    \small\setlength{\tabcolsep}{2pt} 
    \begin{tabular}{llllllll}
\toprule[0.12em]
\textbf{work}   & \textbf{year} & \textbf{area of use} & \textbf{algorithm} & \textbf{database} & \textbf{subjects/session} & \textbf{performance} & \textbf{remarks} \\ \midrule[0.12em]
Golz et al. \cite{golz2007feature} & 2007 & microsleep events & \begin{tabular}[c]{@{}l@{}}support vector machine \\ Learning vector quantization \end{tabular}  & private & 23 subjects & test errors = 9\% & \begin{tabular}[c]{@{}l@{}} Biosignals from EEG and variations in \\ pupil size and eye movements \\ simulated driving, sleep deprivated subjects \end{tabular} \\ \hline

Zhang et al. \cite{Zhang_2017}    & 2017          & \begin{tabular}[c]{@{}l@{}}high-speed train\\operators\end{tabular} & \begin{tabular}[c]{@{}l@{}}SVM builds on FFT\\ features extracted from \\EEG power spectrum\end{tabular} & private           & 10 drivers  & \begin{tabular}[c]{@{}l@{}}precision=90.79\%,\\ sensitivity=86.80\%\end{tabular}            & \begin{tabular}[c]{@{}l@{}}awake/drowsy, wearable \\ BCI model for EEGs, \\ wireless data transmission, \\virtual driving environment\end{tabular} \\ \hline
Jeong et al. \cite{Jeong_Yu_Lee_Lee_2019}    & 2019 & aviation, pilots  & \begin{tabular}[c]{@{}l@{}}deep spatio-temporal \\convolution\end{tabular}            & private  & 9 M, 1 F              & \begin{tabular}[c]{@{}l@{}}2 states Acc = 0.87\\ 5 levels Acc = 0.69\end{tabular} & \begin{tabular}[c]{@{}l@{}}awake/drowsy\\ more fine-grained classification\end{tabular} \\ \hline
Wang et al.  \cite{wang2019spectral}       & 2019          & \begin{tabular}[c]{@{}l@{}}aviation, public\\transportation\end{tabular} & \begin{tabular}[c]{@{}l@{}}handcrafted features\\from EEG, EOG, and\\ facial data\end{tabular} & private           & 16 pilots    & find operational features & aviation headset equipped with sensors        \\ \hline
\begin{tabular}[c]{@{}l@{}}Natnithikarat\\et al. \cite{natnithikarat2019drowsiness}\end{tabular} & 2019          & office employees        & \begin{tabular}[c]{@{}l@{}}linear regression task,\\ PCA, SVM\end{tabular} & private           & \begin{tabular}[c]{@{}l@{}}18 (15 M, 3 F),\\ 1h office work\end{tabular} & \begin{tabular}[c]{@{}l@{}}keystroke, mouse move\\ and self-evaluated KSS \\ correlation to drowsiness\end{tabular} & \begin{tabular}[c]{@{}l@{}}keyboard, mouse events, \\ eye-tracking, EEG+ECG \\ as reference\end{tabular}         \\ \hline

Pham et al. \cite{pham2020wake} & 2020 & microsleep events & \begin{tabular}[c]{@{}l@{}} feature engineering + classifiers \\ deep learnning on raw data \end{tabular}  & private & 19 subjects & 
\begin{tabular}[c]{@{}l@{}}Avg precision = 76\% \\ recall=85\% \end{tabular} & \begin{tabular}[c]{@{}l@{}}Headphones as wearable design\\ noise mitigation and uses EEG+EOG+EMG \end{tabular} \\ \hline

Paulo et al. \cite{Paulo_Pires_Nunes_2020}    & 2021 & driver drowsiness & \begin{tabular}[c]{@{}l@{}}spatio-temporal en-\\coding CNN classifier\end{tabular} & private  & 27 subjects & LOO-CV Acc=75.87\%                 & awake/drowsy   \\ \hline
Chaabene et al. \cite{Chaabene_Bouaziz_Boudaya_Hökelmann_Ammar_Chaari_2021}  & 2021 & driver drowsiness & \begin{tabular}[c]{@{}l@{}}DL-based two-stage,\\ networks\end{tabular} & private  & \begin{tabular}[c]{@{}l@{}}6 M, 8 F,\\ 42 records\end{tabular} & Acc=90.42\%   & awake/drowsy   \\ \hline
Cui et al. \cite{cui2021subject}      & 2021 & driver drowsiness & CNN-LSTM model           & public\footref{fn:eeg}\cite{cao2019multi}  & 11 subjects & Avg Acc=72.97\% (LOO)                & \begin{tabular}[c]{@{}l@{}} awake/drowsy \\ cross subject recognition\end{tabular}  \\ \hline
Cui et al. \cite{cui2022eeg} & 2022 & driver drowsiness & interpretable CNN & private & 11 subjects & Avg Acc=78.35\% (LOO) & \begin{tabular}[c]{@{}l@{}} automatic feature selection from EEG features; \\ cross subject recognition\end{tabular} \\ \hline
Ming et al. \cite{DBLP:journals/tetci/MingWWSL21}    & 2021          & driver drowsiness    & Deep Q-Learning          & private   & 37 subjects & \begin{tabular}[c]{@{}l@{}}use DQN to derive \\ response time from\\ EEG data\end{tabular}     & \begin{tabular}[c]{@{}l@{}}Relate EEG characteristics to \\ response time to indrectly \\ estimate drowsiness states.\end{tabular} \\ \hline
Ramos et al. \cite{Ramos_Maior_Moura_Lins_2022}   & 2022          & \begin{tabular}[c]{@{}l@{}}sectors engaged in\\ safety critical end-\\eavors, oil\&gas \end{tabular} & \begin{tabular}[c]{@{}l@{}}multiple channel EEGs, \\ ensemble machine \\learning\end{tabular} & DROZY\footref{fn:drozy}  & 14 subjects  & \begin{tabular}[c]{@{}l@{}}Accuracy$\geq$ 90\% for\\ specific subjects and \\ dedicated models\end{tabular} & \begin{tabular}[c]{@{}l@{}}awake/drowsy,\\ considered different setups\\ for evaluation.\end{tabular} \\ \hline
Chougule et al. \cite{9997232} & 2022 & microsleep events & \begin{tabular}[c]{@{}l@{}} attention-based method \\ combine STFT+Wavelets \end{tabular}  & MWT dataset \footref{fn:mwt} & 64 (27M,37F) & 
\begin{tabular}[c]{@{}l@{}} train acc=92\% \\ test acc=89.9\% \end{tabular} & One electrode EEG for more user acceptance \\ \hline
Qin et al. \cite{DBLP:journals/ijns/QinNZLJZ23}      & 2023 & driver drowsiness & \begin{tabular}[c]{@{}l@{}}federated learning\\ CNN classifier\end{tabular}       & private  & 11 subjects & \begin{tabular}[c]{@{}l@{}}avg Acc=73.56\%\\ F1-score=73.26\%\\ AUC=78.23\%\end{tabular}    & awake/drowsy   \\ 
\hline
Zhang et al. \cite{zhuang2023connectivity} & 2023 & driver drowsiness & \begin{tabular}[c]{@{}l@{}}Graph Neural network\\ with attention\end{tabular}  & public & \begin{tabular}[c]{@{}l@{}}27 subjects\\62+ sessions \end{tabular} & \begin{tabular}[c]{@{}l@{}}Acc=72.6\% \\ F1 = 70.7\%\end{tabular}  & awake/drowsy\\ \hline
Lee et al. \cite{lee2023lstm} & 2023 & drowsiness detection & LSTM-CNN model  & private & 19 subjects & \begin{tabular}[c]{@{}l@{}}  F1=95\% (4000ms) \\ Acc=85.6\% (500ms) \end{tabular}
 & \begin{tabular}[c]{@{}l@{}}  multistage consciousness \\ (awake, sleep, drowsiness) \\ auditory stimuli and button responses \end{tabular} \\ 
\bottomrule[0.12em]
\end{tabular}
}
\end{table*}

\subsection{ECG-based Drowsiness Detection}

The ECG is another measurement on the skin to record the heartbeat variability. This physiological signal is also often applied for driver drowsiness detection \cite{7479464}. Different to multi-channel EEG signals, ECG does not need to be placed on the scalp, thus can be more suitable for drowsiness detection under more relaxed and natural conditions. Interestingly, the trend goes beyond the development of the algorithms and also affects the design of the sensors. The latest trend shows a shift from traditional, stationary medical placement to a more convenient, and wearable design \cite{Lee_Lee_Shin_2019}.  

Takalokastari et al. \cite{Takalokastari_Jung_Lee_Chung_2011} carried out real-time drowsiness detection utilizing a wireless sensor node connected to a wearable ECG sensor. They build a binary classification method based on extracted features from the ECG signal, which was sampled at 100Hz, to distinguish between awake and drowsy states. The wireless transmission of data facilitated the forwarding of information to a server PC. Notably, the QRS complex in the ECG signal \cite{meyer2006combining} offered valuable features that could aid in the diagnosis of various cardiovascular conditions. The process of drowsiness detection often involves the analysis of R peaks, R-R intervals, the interval between R and S peaks, and the duration of the QRS complex. Another work by Martins et al. \cite{adao2021fatigue} conducted a comprehensive review that examined the latest research on fatigue detection and monitoring using wearable devices. Wearable devices offer a significant advantage by facilitating continuous and long-term monitoring of biomedical signals with comfort and non-intrusiveness. However, the study also identified distinct challenges associated with using wearable devices for fatigue monitoring. 

Later, Shebakova focused on the inter-person variability of the detection scheme. Sherbakova et al. \cite{Sherbakova_Osipova_2015} performed a thorough analysis of ECG signal for driver drowsiness detection. The authors stated that the threshold of drowsiness based purely on individual's heart rate can vary for different people. Possible causes that could affect the threshold of drowsiness include the individual's current postures \cite{achten2003heart} or different cognitive states \cite{weise2013worried} among other factors. Relevant works showed that the heart rate is significantly different when individuals are supine or upright \cite{achten2003heart}. Difference in heart rate is also observable during the sleep phase in a high or low worrier state \cite{weise2013worried}. However, research demonstrated that parameters of HRV change over time depend on the current state (i.e. during wakefulness, drowsiness, and stress). Therefore, the authors suggested using the analysis of three ECG parameters, including heart rate (HR), LF/HF, and the Baevsky stress index \cite{zernov2019cardiometric} as robust indications for drowsiness detection. GPRS data transmission allows the processing and storage of ECG signals on a powerful server.

Ke et al. \cite{Ke_Zulman_Wu_Huang_Thiagarajan_2016} proposed a drowsiness detection system using heartbeat detection from Android-based handheld devices. ECG signal acquired from a sensor is first transferred via Bluetooth to an Android device. The system extracted meaningful information from the ECG signal and indicative features are calculated from the power ratio after applying the hamming window and the Fourier transformation. Data was collected from a male and female test subject both in the awake state as well as in the asleep state. Evaluation results revealed a correlation between the state of drowsiness with a decreasing trend in the ratio (LF/HF). LF band stands for low-frequency component ranging from 0.04$\sim$0.15Hz and HF band stands for high-frequency component ranging from 0.15$\sim$0.4Hz. Each controls certain functionalities in the vegetative nervous system.

A more extended investigation of ECG-based drowsiness detection study was conducted by Fujiwara et al. \cite{Fujiwara_Abe_Kamata_Nakayama_Suzuki_Yamakawa_Hiraoka_Kano_Sumi_Masuda_et}. They proposed a detection algorithm based on heart rate variability (HRV) analysis and validated their method by comparing it with EEG-based sleep assessment. Eight features of heart rate variability are monitored to detect known abnormalities in the signal. During the experimental phase, data were collected from 34 participants in a driving simulator and their sleep stages were labeled by a sleep specialist. Results show that sleepiness was detected in 12 of 13 pre-N1 episodes before sleep onset.  

The work by Lee et al. \cite{Lee_Lee_Shin_2019} introduced a deep learning based approach for drowsiness detection. The authors investigated the robust and deterministic pattern of HRV signals collected from wearable ECG or photolethysmogram (PPG) sensors for driver drowsiness detection. Challenges of using wearable adds additional moving artifacts to collected time series. These motion artefacts can be alleviated. Three types of recurrence plots are generated as input features to a CNN for the binary classification of drowsy and awake state. An experimental dataset was collected under a virtual driving environment to evaluate the proposed measures. 

With a notable surge in motorcycle traffic accidents, frequently leading to serious consequences and a significant loss of life, research on driver drowsiness detection for motorcyclists becomes more relevant \cite{mannering1995statistical}. Motorcyclists are often more vulnerable compared to car drivers in case of accidents \cite{DBLP:journals/jaihc/Ospina-MateusJL21}. To address this concern, Fahrurrasyid et al. \cite{Fahrurrasyid_Hapsari_Meisaroh_Mutiara_2022} introduced an innovative solution: a smart helmet integrated with various sensors. These sensors monitor psychological signals, including heartbeats, alongside a GPS module, GSM module, and alert push notifications. The study's experiment, involving 10 participants, demonstrated the helmet's capability to identify drowsiness and send alerts when the heart-rate drops below 60 bpm. This data is accessible in real-time, while the helmet also employs the Google Maps application to track the precise location of the incident.

Latest interesting work by Heydari et al. \cite{heydari2022detection} introduced a technique to identify driver drowsiness by monitoring the pulse rate variability (PRV) measured on a finger. They analyzed finger pulse data which are derived from PPG signals, focusing on features within the pulse rate variability that exhibit notable changes during drowsiness. Findings reveal that the variability values, along with their averages, increased before the onset of sleepiness. Additionally, it was observed that the standard deviation of all peak-to-peak intervals notably decreases during drowsiness. Also, an increase in the values of the Root Mean Square of Successive Differences (RMSSD) is observed during the drowsiness stage. The authors suggested a purely conceptual design to integrate PPG sensors into a steering wheel to detect the driver's finger pulse rate, offering a viable and non-invasive means for detecting driver drowsiness.

Due to the subtlety of microsleeps, these events are usually recorded using EEG or EOG signals from subtle eye muscle movements. Towards that, Lenis et al. \cite{lenis2016detection} proposed a work investigating MSEs during a car driving simulation using ECG features. In this work, morphological and rhythmical features before and after a MSE are extracted from the ECG signals and analyzed towards baseline. The findings suggested that detecting (or predicting) MSE solely based on the ECG is not feasible. However, when MSE is present, noticeable differences in both the rhythmic and morphological features were observed compared to those calculated for the reference signal in the absence of sleepiness.

Table \ref{tab:ECG} summarizes the investigated works using ECG signals for drowsiness detection in various application scenarios. Sensing modality covers both wired and wireless wearable devices targeting non-intrusive applications. References to the public databases are provided in the footnote if available. From Table \ref{tab:ECG}, we notice that beyond the development of algorithmic choices, i.e., starting from a more heuristic pattern generation to more advanced recurrent and CNN-based methods, the trend also goes to the evaluation of larger groups with more subjects and in the design choices of more flexible and wearable sensors introduce their own individual pros and cons.

\begin{table*}[]
\caption{summarizes recent works performing drowsiness detection based on ECG signals.}
\label{tab:ECG}
    \centering
    \resizebox{\textwidth}{!}{%
    \small\setlength{\tabcolsep}{2pt} 
    \begin{tabular}{llllllll}
\toprule[0.12em]
\textbf{work}        & \textbf{year} & \textbf{area of use}  & \textbf{algorithm}             & \textbf{database} & \textbf{subjects/session}         & \textbf{performance}          & \textbf{remarks}         \\ \midrule[0.12em]
Takalokastari et al. \cite{Takalokastari_Jung_Lee_Chung_2011} & 2011          & driver drowsiness     & \begin{tabular}[c]{@{}l@{}}heart rate variability\\ handcrafted, heuristic\end{tabular} & private           & -              & find operation thresholds     & wireless sensor node, wearable ECG           \\ \hline
Sherbakova et al. \cite{Sherbakova_Osipova_2015}   & 2015          & public transportation & \begin{tabular}[c]{@{}l@{}}heart rate variability\\ handcrafted, heuristic\end{tabular} & private           & -              & find operation thresholds     & \begin{tabular}[c]{@{}l@{}}portable device for 1 lead ECG,\\ wireless data transmission\end{tabular} \\ \hline
Ke et al.   \cite{Ke_Zulman_Wu_Huang_Thiagarajan_2016}         & 2016          & driver drowsiness     & \begin{tabular}[c]{@{}l@{}}HRV, FFT\\ handcrafted, heuristic\end{tabular}               & private           & \begin{tabular}[c]{@{}l@{}}1 M, 1 F,\\ 120 minutes\end{tabular} & find operation thresholds     & handheld devices, wireless transmission     \\ \hline
Lenis et al. \cite{lenis2016detection} & 2016 & microsleep event & \begin{tabular}[c]{@{}l@{}} ECG features before \\ and after MSE\end{tabular}  & private & \begin{tabular}[c]{@{}l@{}} 7 subjects; \\ 14 records \end{tabular} & 
- & \begin{tabular}[c]{@{}l@{}} finding significant changes \\ in heart rate variability \\ around MSE \end{tabular} \\ \hline
Fujiwara et al. \cite{Fujiwara_Abe_Kamata_Nakayama_Suzuki_Yamakawa_Hiraoka_Kano_Sumi_Masuda_et}     & 2018          & driver drowsiness     & \begin{tabular}[c]{@{}l@{}}heart rate variability\\ handcrafted features\end{tabular}   & private           & 34 subjects    & \begin{tabular}[c]{@{}l@{}}12 out of 13 pre-N1\\ episodes prior sleep\\ onsets detected\\ FP=1.7 times per hour\end{tabular} & \begin{tabular}[c]{@{}l@{}}data labeled by sleep specialist\\ detection of sleep onsets\end{tabular} \\ \hline
Lee et al.   \cite{Lee_Lee_Shin_2019}        & 2019          & driver drowsiness     & \begin{tabular}[c]{@{}l@{}}Recurrent methods\\ CNN classification\end{tabular}          & private           & \begin{tabular}[c]{@{}l@{}}6 subjects,\\ 22 recordings\end{tabular}     & \begin{tabular}[c]{@{}l@{}}ReLU-RP CNN:\\ ECG Acc = 70\% \\ PPG Acc = 64\%\end{tabular}   & wearable ECG + PPG sensors  \\ \hline
Fahrurrasyid et al. \cite{Fahrurrasyid_Hapsari_Meisaroh_Mutiara_2022}  & 2022          & motorcycle driver     & \begin{tabular}[c]{@{}l@{}}heart rate variability\\ handcrafted, heuristic\end{tabular} & private           & 10 subjects    & find operation thresholds     & a smart helmet equipped with sensors        \\ \hline
Heydari et al. \cite{heydari2022detection}  & 2022          & driver drowsiness    & \begin{tabular}[c]{@{}l@{}} pulse rate variability \\ of a finger from PPG\end{tabular}  & private           & 10 subjects  & relevant features selection        & \begin{tabular}[c]{@{}l@{}} awake/drowsy, heuristic features from time, \\ frequency domain, and nonlinear analysis \end{tabular}                \\ \hline
Hasan et al. \cite{hasan2024validation} & 2024 & driver drowsiness& \begin{tabular}[c]{@{}l@{}} explainable ML in \\multimodal system \end{tabular}  &  private & 35 subjects & 
\begin{tabular}[c]{@{}l@{}} sensitivity=70.3\% \\ specificity=82.2\% \\ Acc=80.1\% \end{tabular} & \begin{tabular}[c]{@{}l@{}} validation techniques for black box model \\ combining EEG+EOG+ECG signals \end{tabular} \\
\bottomrule[0.12em]
\end{tabular}
}
\end{table*}

\subsection{Vision-based Drowsiness Detection}
Vision-based drowsiness detection is intended to be non-invasive and non-intrusive. Unlike physiological signals, this method does not require close contact with the subject. It operates remotely and does not necessitate physical attachment or direct interaction with the individual being monitored. Most relevant features for vision-based drowsiness detection are focused on the facial attributes\cite{garcia2012vision,Bakheet_Al-Hamadi_2021}, such as eye blinking, eye aspect ratio or facial expressions\cite{deng2019real,Vijay_Vinayak_Nunna_Natarajan_2021,ahmed2021intelligent} such as yawning or mouth opening which indicates the level of drowsiness. In contrast to the works developed based on physiological signals from previous sub-sections, it exists more official databases for the development and evaluation of vision-based drowsiness detection schemes.

Earlier work by Garcia et al. \cite{garcia2012vision} proposed a vision-based drowsiness detector under real driving conditions. An infrared camera is placed in front to capture the driver's face and to obtain drowsiness clues from their eyes closure. Three stages of processing include face and eye detection, pupil position detection, and illumination adaptation. Finally, the PERCLOS features are extracted to relate them to the drowsiness state. An outdoor database of several experiments over 25 driving hours was generated as the evaluation dataset. Results of the binary classification for awake and fatigue states showed a specificity, sensitivity, and recall of 92.23\%, 79.84\%, and 90.68\% respectively.

In this research \cite{yu2017representation}, Yu et al. introduced an innovative approach for drowsiness detection, utilizing three main steps for simultaneous representation learning, scene understanding, and feature fusion. They extracted and learned spatio-temporal representations from consecutive frames and employed scene conditional understanding and fusion techniques to enhance the accuracy of drowsiness detection. To evaluate their method's performance, they tested it on the NTHU-DDD dataset. Results showed a validation accuracy of 88\% and an F1-score of 0.712. However, the limitation of the proposed model is its generalizability. Since it is trained on NTHU-DDD dataset, it may not be directly applicable to scenarios that deviate from the trained conditions.

Deng et al. \cite{deng2019real} proposed a system called DriCare which unobtrusively detects the driver's fatigue status clues, such as yawning, blinking, and duration of eye closure, based on video images. They introduced a face-tracking algorithm to improve the tracking accuracy and designed a new detection scheme for facial regions based on 68 facial landmarks which are leveraged to access the driver's state. By fusion features of the eyes and mouth, DriCare achieved an accuracy of 92\% on the YawDD database.

To mitigate the problem of changing illumination under real driving conditions in a car, Bakheet et al. \cite{Bakheet_Al-Hamadi_2021} proposed an improved histogram of oriented gradients (HOG) features combined with a naive Bayesian classification to detect driver drowsiness. The experimental outcomes on the publicly accessible NTHU-DDD dataset demonstrated that the proposed framework has the potential to compete strongly with several SOTA baselines. Results showed an average accuracy of 85.62\%. However, the model could have the same shortcomings of missing generalizability.

Vijay et al. \cite{Vijay_Vinayak_Nunna_Natarajan_2021} presented a vision-based, two-stage pipeline for real-time driver drowsiness detection using Facial Action Units (FAUs). FAUs can represent facial expression-related movements in facial muscle groups. In the first stage, they employed CNN for detecting FAUs. The second stage utilized an Extreme Gradient Boosting (XGBoost) classifier for drowsiness detection. To model user-specific behavior, individual classifiers were trained. This approach achieved high accuracy in real-time using only a small amount of data and short training time.

Liu's work \cite{LIU2021103901} focused on the drowsiness detection of crane operators by using deep neural networks leveraging both spatial features and temporal features.  They combined the spatial feature extraction with CNN and temporal feature extraction with a Long-short-term-memory (LSTM) network. The authors collected facial videos from licensed crane operators under simulated crane operation scenarios and created a large and public fatigue dataset especially tailored for crane operators. They trained their model on three public vehicle driver datasets, NTHU-DDD, UTA-RLDD, and YawnDD, with human-verified labels at the frame and minute segment levels.

Ahmed et al. \cite{ahmed2021intelligent} proposed an approach combining visual features from both eyes and mouth regions extracted from two separate CNNs for driver drowsiness detection. The weights of each stream are further trained on a single-layer perceptron to output the final prediction of drowsiness or non-drowsiness detection. The strength of the ensemble structure is demonstrated over single-stream processing using one of the two face areas. This model is evaluated on the NTHU-DDD video dataset \cite{weng2017driver} with an accuracy of 97.1\,\% and showed robustness over variations in pose and illumination. More recent multi-stream classification networks combining both spatial and spatio-temporal features is proposed by Pandey et al. \cite{PANDEY2023105759}.

Krishna et al. \cite{krishna2022vision} proposed to build a driver drowsiness detection framework by fusion object detection and using global attention. Thus, the authors leveraged vision transformers and YoloV5 detectors in their proposed framework. This work aims to capture more complicated driver behaviour features from images compared to current CNN-based methods in this field. The framework is evaluated on the public dataset UTA-RLDD and further validated on a custom dataset of 39 participants collected under various light conditions. On both datasets, the proposed method showcased promising results in terms of high accuracy. The authors claimed the significance of their proposed framework for practical applications in smart transportation systems. 

Tamanani et al. \cite{tamanani2021estimation} proposed a new driver's vigilance detection system based on deep learning methods on facial region diagnosis using the Haar-cascade method and CNN for classification. Evaluation is performed on the UTA-RLDD dataset with five-fold cross-validation. Results showed an accuracy of 96.8\% which is higher than most previously reported algorithms. Another customized dataset with 10 subjects under different light conditions was collected to evaluate the generalizability of the proposed method. 

In addition to the detection of eye blinking and yawning in a visual image, Khan et al. \cite{DBLP:journals/access/KhanNKHR23} further considered another visual feature as an indicator of 'distraction' by detecting the driver looking sideways for a certain duration of times, i.e. more than 3 seconds. This feature is calculated by determining the Euclidean distances from both ears to nose tip, which builds a triangle, and the difference of both distances is related to side looking. Side looking face will cause an increase in this difference measure compared to frontal view. To evaluate the proposed method, experiments were conducted on a self-collected dataset containing 50 subjects. Results showed a precision, recall, and F1-score of 0.89, 0.98, and 0.93 respectively.

Most vision-based detection methods primarily focus on frontal faces and struggle to handle various head poses encountered in real driving scenarios. Chen et al. \cite{9922476} dealt with the challenge by proposing a network to accurately detect driver drowsiness from various viewing angles combining transfer learning and population-based sampling strategy (TLPSN). The population-based sampling strategy is adopted to curate a new training set from data captured in a driver-in-the-loop platform. The results demonstrated that the proposed method has strong robustness to the variation of pose while maintaining high accuracy. In addition, transfer learning significantly improves the generalizability of the model.

Lu et al. \cite{10159554} recently introduced a novel network also designed for detecting driver yawning across arbitrary poses in video. The network comprises three key components: a Geometric-based Key-frame Selection Module (GK-Module), a Face Frontalization with Warp Attention Module (FF-Module), and a dual-channel classifier for Head Pose \& Facial Action Fusion Module (HF-Module). Extensive experiments demonstrate that the proposed JHPFA-Net achieves SOTA performance compared to several representative methods on the public YawDD benchmark. Moreover, it exhibits excellent performance in real-time applications.

Table \ref{tab:Camera} summarizes the works using images or video data for drowsiness detection in various application scenarios. Vision-based drowsiness detection approaches are non-intrusive and remote, focusing mostly on facial clues, eye blink rate, or head positions. References to the public databases used as benchmarks are provided in the footnote. From Table \ref{tab:Camera}, we noticed that there exist more public databases for the development and evaluation of vision-based drowsiness detection schemes compared to physiological signals. A similar trend from traditional machine learning to deep learning-based methods for drowsiness detection can be observed over time. Sequence models and attention-based vision transformers represent the latest advancements in deep learning-based drowsiness detection schemes.

\begin{table*}[]
\caption{summarizes recent works performing drowsiness detection based on vision.}
\label{tab:Camera}
    \centering
    \resizebox{\textwidth}{!}{%
    \small\setlength{\tabcolsep}{2pt} 
    \begin{tabular}{llllllll}
\toprule[0.12em]
\textbf{work}  & \textbf{year} & \textbf{area of use}  & \textbf{algorithm}            & \textbf{database} & \textbf{subjects/session}           & \textbf{performance} & \textbf{remarks}             \\ \midrule[0.12em]

Garcia et al. \cite{garcia2012vision}  & 2012          & driver drowsiness     & \begin{tabular}[c]{@{}l@{}}face, eye, pupil detection\\ illumination adaptation\\ heuristic\end{tabular}  & private           & \begin{tabular}[c]{@{}l@{}}10 subjects, 30 h\\ driving, 1296 awake\\min., 504 fatigue min.\end{tabular} & \begin{tabular}[c]{@{}l@{}}Specificity=92.23\%\\ Sensitivity=79.84\%\\ Recall=90.68\%\end{tabular} & under real drive conditions  \\ \hline
Yu et al.  \cite{yu2017representation}    & 2017          & driver drowsiness     & \begin{tabular}[c]{@{}l@{}}representation learning, \\ scene understanding,\\ feature fusion\end{tabular} & NTHU-DDD \footref{fn:nthu} & 36 subjects      & \begin{tabular}[c]{@{}l@{}}validation Acc=88\%\\ F1-score = 0.712\end{tabular}  & \begin{tabular}[c]{@{}l@{}}feature sparsity in fusion model,\\ framework may not generalize\end{tabular} \\ \hline
Deng et al. \cite{deng2019real} & 2019 & driver drowsiness & \begin{tabular}[c]{@{}l@{}}imroved face tracking, \\ features from face regions\end{tabular} & YawDD\footref{fn:yawdd} & 107 subjects & Acc = 92\% & \begin{tabular}[c]{@{}l@{}} real-time system,\\
applicable to different circumstances\end{tabular} \\ \hline
Bakheet et al. \cite{Bakheet_Al-Hamadi_2021} & 2021          & driver drowsiness     & \begin{tabular}[c]{@{}l@{}}HOG feature\\ Naive Bayesian Classifier\end{tabular}        & NTHU-DDD \footref{fn:nthu}         & 36 subjects      & Acc = 85.62\%     & \begin{tabular}[c]{@{}l@{}}limitation of generalizability,\\ need more diverse datasets\end{tabular}     \\ \hline
Vijay et al. \cite{Vijay_Vinayak_Nunna_Natarajan_2021}  & 2021          & driver drowsiness     & \begin{tabular}[c]{@{}l@{}}CNN for Facial Action\\ Units, Extreme Gradient \\ Boosting Classifier\end{tabular}      & NTHU-DDD \footref{fn:nthu}          & 36 subjects      & Acc =96\%         & subject-specific classification     \\ \hline
Liu et al.    \cite{LIU2021103901}      & 2021          & crane operator  & LSTM + CNN         & \begin{tabular}[c]{@{}l@{}}NTHU\footref{fn:nthu}, UTA\footref{fn:uta},\\  YawnDD\footref{fn:yawdd}, Custom\end{tabular} & - & - & \begin{tabular}[c]{@{}l@{}}simulated crane operation,\\ made their database public\end{tabular}        \\ \hline
Ahmed et al. \cite{ahmed2021intelligent}  & 2021          & driver drowsiness     & two streams CNNs              & NTHU-DDD \footref{fn:nthu}         & 36 subjects      & \begin{tabular}[c]{@{}l@{}}Evaluation dataset\\ Acc=97.1\%\end{tabular}         & \begin{tabular}[c]{@{}l@{}}robust over variations\\ in pose and illumination\end{tabular}                \\ \hline
Mou et al. \cite{mou2021isotropic} & 2021 & driver drowsiness & \begin{tabular}[c]{@{}l@{}} IsoSSL-MoCo \end{tabular}  &  \begin{tabular}[c]{@{}l@{}} NTHU-DDD\footref{fn:nthu}  \\ YawDD\footref{fn:yawdd}  \end{tabular} & \begin{tabular}[c]{@{}l@{}} 36 subjects \\ 107 subjects\end{tabular} & 
\begin{tabular}[c]{@{}l@{}} Acc=93.71\% \\ Acc=98.65\% \end{tabular} & \begin{tabular}[c]{@{}l@{}} pretrain on MRL dataset\footref{fn:mrl}  + NTHU-DDD\footref{fn:nthu}  \\ leveraging self-supervised learning \end{tabular} \\ \hline
Krishna et al. \cite{krishna2022vision}  & 2022          & smart transportation & vision transformers + Yolov5                & \begin{tabular}[c]{@{}l@{}}UTA-RLDD\footref{fn:uta}\\ Custom\end{tabular} & \begin{tabular}[c]{@{}l@{}}60 subjects,\\ 39 subjects\end{tabular}         & \begin{tabular}[c]{@{}l@{}}train Acc = 96.2\%\\ valid Acc = 97.4\%\\ custom Acc = 95.5\%\end{tabular} & \begin{tabular}[c]{@{}l@{}}attention-based model + object detection\\ vigilent/drowsy detection\end{tabular} \\ \hline

Chen et al. \cite{9922476} & 2022 & driver drowsiness & \begin{tabular}[c]{@{}l@{}}multiple viewing angle \\ sampling strategy for \\ data augmentation \end{tabular}  & private & 5 subjects & \begin{tabular}[c]{@{}l@{}}Acc = 97.5\% \\ F1 = 97.5\% \end{tabular}  & \begin{tabular}[c]{@{}l@{}}simulated driving, \\ sleep deprivated subjects \\ small-scale dataset  \end{tabular} \\ \hline
Tamanani et al. \cite{tamanani2021estimation}    & 2023          & public transportation           & \begin{tabular}[c]{@{}l@{}}facial region diagnosis \\ (with Haar-cascade),\\ CNN classification\end{tabular} & \begin{tabular}[c]{@{}l@{}}UTA-RLDD\footref{fn:uta} ,\\ private\end{tabular} & \begin{tabular}[c]{@{}l@{}}60 subjects,\\ 10 subjects\end{tabular}    & \begin{tabular}[c]{@{}l@{}}avg Acc = 0.918\\ precision = 0.928\\ recall = 0.920\\ F1-score = 0.920\end{tabular} & \begin{tabular}[c]{@{}l@{}}authors evaluated their method\\ on a customized dataset\end{tabular}       \\ \hline
Khan et al. \cite{DBLP:journals/access/KhanNKHR23}    & 2023          & public transportation & \begin{tabular}[c]{@{}l@{}}handcrafted features, \\ heuristics\end{tabular}            & private           & \begin{tabular}[c]{@{}l@{}}50 subjects \\ (33 M, 17 F)\end{tabular}                & \begin{tabular}[c]{@{}l@{}}Precision = 0.89\\ Recall = 0.98\\ F1-score =0.93\end{tabular}          & \begin{tabular}[c]{@{}l@{}}IoT-based Non-Intrusive\\ to enhance Road Safety\end{tabular}                 \\ \hline
Pandey et al. \cite{PANDEY2023105759} & 2023 & driver drowsiness & \begin{tabular}[c]{@{}l@{}} mutlistream classifcation \\ YoLov3 + LSTM \end{tabular}  & UTA-RLDD\footref{fn:uta} & 60 subjects & 
\begin{tabular}[c]{@{}l@{}} Acc=97.5\% \end{tabular} & \begin{tabular}[c]{@{}l@{}} Spatio-temporal feature; \\ TransGAN; YOLOv3 \\Temporal feature; LSTM \end{tabular} \\ \hline

Lu et al \cite{10159554} & 2023 &  driver drowsiness & DL with 3 modules  & YawDD\footref{fn:yawdd}  & 107 subjects & \begin{tabular}[c]{@{}l@{}}Acc=86.7\% \\ Precision = 91\% \\ F1-score =88.8\%\end{tabular}    & targets arbitrary head poses \\\bottomrule[0.12em]
\end{tabular}
}
\end{table*}
\section{Widely Used Databases}
\label{sec:database}

This section summarizes popular databases used for drowsiness detection based on visual, mutli-channel EEG, and ECG signals. This discussion stressed on available databases and put less focus on small-scale private databases. Table \ref{tab:databases} contains an overview of these publicly available databases for drowsiness detection research. A link to the individual database is given in the footnote for easier access. In Table \ref{tab:databases}, we target this tabular representation from various aspects, including the database name, year of publication (chronologically ordered), detection modality, number of subjects/sessions included, its label annotation, application area, and conclude with specific remarks. The majority of the cited databases here are used for driver drowsiness detection but can be extended to general drowsiness detection tasks because of the diversity of the capture environments.

The University of Texas at Arlington created the Real-Life Drowsiness Dataset\footnote{UTA-RLDD: https://sites.google.com/view/utarldd/home\label{fn:uta}} (UTA-RLDD) \cite{Ghoddoosian_2019_CVPR_Workshops} targeting the task of multi-stage drowsiness detection. The dataset contains both easily visible cases and subtle cases where the drowsiness level is at an early stage and the detection is strongly related to subtle micro-expressions. The creators claimed that the UTA-RLDD dataset is the largest to date realistic drowsiness dataset. It consists of around 30 hours of RGB videos of 60 healthy participants, where each participant collected three different classes including alertness, low vigilance, and drowsiness. It contains variations in gender, ethnicity, age, acquisition device, and also accessories like glasses, facial hair and different viewing angles in different real-life environments and backgrounds. 
The position of the acquisition device is placed such that both eyes are visible and the device is within one arm length away from the subject. This database is often used to evaluate algorithms for driver's drowsiness detection with vision-based approaches\cite{tamanani2021estimation}. 

The academic NTHU-DDD dataset\footnote{NTHU-DDD: http://cv.cs.nthu.edu.tw/php/callforpaper/datasets/DDD/\label{fn:nthu}}   \cite{weng2017driver} was collected by NTHU Computer Vision Lab at National Tsing Hua University. This is another video dataset for detecting driver's drowsiness. They used active infrared illumination to alleviate the poor illumination issue. All videos were captured by a stand-alone surveillance camera D-Link DCS-932L with a resolution of 640x480 pixels. The capture device is placed on the top left to emulate the position in the car without blocking the driver's view. Each subject recorded two different sessions including day-time and night-time sessions. Subjects are asked to perform actions indicating different drowsiness levels while playing a plain driving game, such as normal driving, slow blink rate, yawning, falling asleep and bursting out laughing. The dataset includes variations in illumination, scenarios, skin color, gender, age and also differences in hairstyles, clothing and glasses/sunglasses to cover the most realistic driving scenarios. Compared to UTA-RLDD, although this database was also recorded under indoor settings, its applications are much more limited, because the use-case collected here is more controlled.

The YawDD dataset\footnote{YawDD: https://traces.cs.umass.edu/index.php/Mmsys/Mmsys\label{fn:yawdd}} \cite{abtahi2014yawdd} is recorded by an in-car camera capturing the driver's facial characteristics in an actual car during talking, singing, being silent, and yawning. It contains two different camera installations: (1) under the front mirror and (2) on the driver's dashboard. In total 322 RGB videos were recorded with large variations in illumination, gender, ethnicity, and with and without glasses/sunglasses. Additional 29 videos, one for each subject, are added under second setup containing all sequences of performed actions. Compared to UTA-RLDD or NTHU-DDD this database is more realistic in the data acquisition, as the videos are collected in a real vehicle with individuals sitting behind the wheel. This allows the design of drowsiness detection algorithm under more realistic scenarios. The challenge in this database accounts for the placement of the camera under the mirror. Under this viewing angle face detection is much harder.

Media Research Lab from the Technical University of Ostrava proposed the MRL Eye Dataset\footnote{Project page MRL Eye Dataset: http://mrl.cs.vsb.cz/eyedataset\label{fn:mrl}} \cite{Fusek2018433} which could be used to detect eye blinking and relate this to drowsiness detection by combining eye blinking rate to drowsiness levels. This database was not originally designed for drowsiness detection. But it can be leveraged to detect visual features of the human eyes related to drowsiness. It is a large-scale dataset of human eye images and consists of 84,898 images from 37 individuals, while 33 male and 4 female subjects are included. This dataset further modulated variations in capture devices, capture spectra (RGB + IR), lightning conditions, image resolutions and eye openness levels. Unlike previous databases, this database restricts the area of faces to only the eye regions making the detection of subtle facial micro-expressions impossible.

The FatigueView dataset\footnote{Project page FatigueView Dataset: https://fatigueview.github.io/\#/\label{fn:fatigue}}, introduced by Yang et al. \cite{DBLP:journals/tits/YangYLS23}, is a large-scaled dataset of videos designed for driver's drowsiness detection. This dataset boasts practicality, diversity, and a wide range of environments. It encompasses images captured by both RGB and infrared cameras, positioned in five different angles. The dataset includes genuine instances of drowsy driving and displays various visual indications of different drowsiness levels. Notably, the dataset features a substantial 17,403 instances of yawning, significantly surpassing current widely used datasets. The authors evaluated the dataset using SOTA algorithms, therefore establishing numerous baseline results that can guide future algorithm advancements. Unlike the NTHU-DDD, this database is collected in a single office environment by sitting the participants on an upholstered office chair. While the acquisition environment is not diverse enough compared to NTHU-DDD or UTA-RLDD, it covers a much broader viewing angle.

The National Cheng Kung University Driver Drowsiness Dataset (NCKUDD) \cite{chiou2019driver} consists of videos from a total of 25 participants captured during a normal driving condition. The camera is placed in front of the driver to unobtrusively capture the driver's facial expression. Recordings contain both daylight and dark environments and include normal, sleepy, distracted, talking/eating scenarios while driving, talking on the phone while driving, and other abnormal driving patterns. Compared to other vision-based databases so far, this database may contain footage of participants maneuvering the car for real. While this fact might be considered advantages, it may include situations that endanger the driver's safety.

A novel database is introduced by Martin et al. \cite{martin2019drive} called the Driver\&Act dataset\footnote{Project page Drive\&Act: https://driveandact.com/\label{fn:driveAct}}, uniquely designated for fine-grained recognition of driver behavior in autonomous vehicles. This comprehensive dataset encompasses more than 9.6 million frames spanning 12 hours, capturing individuals engaged in distracting actions during both manual and automated driving scenarios. The dataset utilizes a capture device that integrates RGB, IR, depth, and 3D body pose data from six different perspectives. Among its objectives, this database particularly aims to excel in the identification of intricate actions and features a multi-modal approach to activity recognition. As opposed to other drowsiness detection databases, this database specifically target the use-case of autonomous driving and was not originally addressed for drowsiness detection only.

Ortega et al. \cite{ortega2020dmd} proposed DMD, which is another large-scale multi-modal driver monitoring dataset\footnote{Project page DMD dataset: https://dmd.vicomtech.org/\label{fn:dmd}} for attention and alertness analysis. This is an extensive dataset with 37 drivers which includes real and simulated driving scenarios. Targeted tasks contain levels of distraction, gaze allocation, drowsiness detection, hands-wheel interaction and context data. It contains in total 41 hours of RGB, depth and infrared videos from 3 cameras capturing face, body, and hands. Compared to other existing similar datasets, the authors motivated their proposed database to be more extensive, diverse, and multi-purpose. Unlike UTA-RLDD, NTHU-DDD or YawDD, this database does not consider drowsiness as the main focus of distraction and considers much more diverse activities causing the drivers to shift their attention from the road, such as talking on the phone, playing with the car interface, or typing into a car navigation system.  

Recently, Yilmaz et al. \cite{yilmaz2022sust} introduced a novel dataset, termed SUST-DDD\footnote{SUST-DDD: https://www.kaggle.com/datasets/esrakavalci/sust-ddd\label{fn:sust}}, aimed at driver drowsiness detection. This dataset is meticulously curated and benchmarked using a range of DL techniques to effectively predict driver drowsiness. The dataset compilation involved 19 participants who were instructed to record themselves while driving using their personal cell phones positioned in front of the driver's seat. To closely replicate real-world driving scenarios, participants were required to operate their own vehicles and utilize their individual phones. Notably, this dataset encapsulates genuine driving situations encompassing diverse lighting conditions, distinct video sizes, and varying resolutions due to each participant's unique phone specifications. This database is most similar to YawDD and NCKUDD, however with the difference that both YawDD and SUST-DDD use user-specific cell phones, while only SUST-DDD includes real driving scenarios. The difference between NCKUDD and SUST-DDD is that NCKUDD uses a camera installed in front of the driver, while in SUST-DDD custom cell phones are used as recording devices within custom vehicles to simulate a more natural driving experience.

The ULg Multimodality Drowsiness Database\footnote{Project page DROZY: http://www.drozy.ulg.ac.be/\label{fn:drozy}}  (DROZY) \cite{massoz2016ulg} is a database which is not only based on vision capture to detect drowsiness. Here, a total of 14 subjects (3 males, 11 females) have participated in the data collection process. Experiments were conducted in a quite and isolated laboratory environment. Participants have no sleeping disorders and a sleep diary was kept individually. Tea and coffee were avoided during the acquisition process to keep the subjects really drowsy instead of faking it. In this database, four different electrical bio-signals such as electroencephalogram (EEG), electroculography (EOG), electrocardiogram (EKG) and electromyogram (EMG) were acquired. Similar to UTA-RLDD, this database was also not specifically designated to target driver's drowsiness detection but includes much more sensory inputs especially measuring physiological entities compared to UTA-RLDD.

Last but not least, another EEG-based database is proposed by Zhao et al. \cite{cao2019multi} in Multichannel EEG\footref{fn:eeg}. They collected 32-channel EEG data during a sustained-attention driving task for drivers' drowsiness detection. In Table \ref{tab:databases}, we referred this database to the Multi-channel EEG recordings. It consisted of more than 62 sessions for 27 subjects. Each session includes a 90-minute sustained-attention task in which an immersive driving scenario is simulated. The participants were asked to drive on a four line highway and keep the car in the center of the lane. Random car drifts are induced expecting the participants to respond accordingly. The driver's drowsiness is inferred from the required response time in this lane-keeping task. This dataset is specifically collected for targeting the driver's drowsiness detection task.

In this section, we have seen a wide range of databases that motivate research on drowsiness detection. Both visual and biosignal-based databases are investigated here. While most databases, such as NTHU-DDD, YawDD, FatigueView, NCKUDD, Drive \& Act, DMD, and SUST-DDD, are intended for driver drowsiness detection, other databases, such as UTA-RLDD, MRL Eye Dataset, and ULg Dataset, are targeting the more general drowsiness detection tasks. Especially the UTA-RLDD database is collected under unconstrained real-world indoor scenarios, which can be used to develop drowsiness detection applications more general for smart environments. But also NHU-DDD collected under indoor scenarios while sitting can be extended to build general drowsiness detection applications for smart environments as well.

\begin{sidewaystable*}[]
\caption{summarizes databases used for drowsiness detection. Most of them are used for driver drowsiness detection but can be easily extended to general drowsiness detection framework for smart environments, as they are collected under various indoor scenarios.}
\label{tab:databases}
    \centering
    \newcolumntype{R}{>{\raggedright \arraybackslash} X}
    \newcolumntype{S}{>{\centering \arraybackslash} X}
    \newcolumntype{T}{>{\raggedleft \arraybackslash} X}
    \resizebox{\textwidth}{!}{%
    \small\setlength{\tabcolsep}{1pt} 
    \begin{tabularx}{\linewidth} {>{\setlength\hsize{.09\hsize}}R >{\setlength\hsize{.04\hsize}}R 
    >{\setlength\hsize{.13\hsize}}R >{\setlength\hsize{.16\hsize}}R >{\setlength\hsize{.10\hsize}}R >{\setlength\hsize{.18\hsize}}R 
    >{\setlength\hsize{.08\hsize}}R > {\setlength\hsize{.18\hsize}}R >{\setlength\hsize{.20\hsize}}R >{\setlength\hsize{.24\hsize}}R}  
  \toprule[0.12em]
\textbf{database} & \textbf{year} & \textbf{measrung technique} & \textbf{attributes} & \textbf{sensing modality} & \textbf{area of application} & \textbf{subjects} & \textbf{sessions}       & \textbf{classes}     & \textbf{specific remarks} \\ \midrule[0.12em]

YawDD\footref{fn:yawdd} \cite{abtahi2014yawdd}                                                       & 2014          & vision-based                                                            & \begin{tabular}[c]{@{}l@{}}yawning,\\ open mouth\end{tabular}                    & remote                    & driver drowsiness                                                                               & 29                   & 342 RGB videos                                                                  & normal, talking/singing, yawing, used in \cite{LIU2021103901, deng2019real, weng2017driver}                                                 & recorded in an actual parked car                                                                                                              \\ \hline
NTHU-DDD\footref{fn:nthu} \cite{weng2017driver}                                                    & 2017          & vision-based                                                            & \begin{tabular}[c]{@{}l@{}}head pose,\\ yawning,\\ eye closure\end{tabular}      & remote                    & \begin{tabular}[c]{@{}l@{}}driver drowsiness,\\ while sitting on a \\ chair indoor\end{tabular} & 22                   & 360 RGB videos                                                                  & drowsiness and non-drowsiness, used in \cite{ahmed2021intelligent,yu2017representation,Bakheet_Al-Hamadi_2021,Vijay_Vinayak_Nunna_Natarajan_2021}                                                       & controlled indoor scenarios, under simulations, contains variations for most realistic driving scenarios                                                           \\ \hline
MRL Eye Dataset\footref{fn:mrl} \cite{Fusek2018433} & 2018          & vision-based                                                            & \begin{tabular}[c]{@{}l@{}}eye blink,\\ eye openess\end{tabular}                 & remote                    & general drowsiness                                                                              & 37                   & \begin{tabular}[c]{@{}l@{}}85000 different\\ eye regions\end{tabular}           & \begin{tabular}[c]{@{}l@{}}driver drowsiness, \\ gaze direction, \\ used in \cite{el2023machine,saurav2022real}  \end{tabular}       & contains only eye regions not faces, impossible to detect facial expressions                         \\ \hline

UTA-RLDD\footref{fn:uta} \cite{Ghoddoosian_2019_CVPR_Workshops}                                                    & 2019          & vision-based                                                            & facial expressions extreme to subtle                                             & remote                    & \begin{tabular}[c]{@{}l@{}}general drowsiness\\ driver drowsiness\end{tabular}                  & 60                   & 180 RGB videos                                                                  & alert, low vigilance, drowsiness, used in \cite{tamanani2021estimation}                                                  & contains variations in gender, ethnicity, age, acquisition device and accessories, can be used for general drowsiness detection \\ \hline
\begin{tabular}[c]{@{}l@{}}NCK-\\ UDD \cite{chiou2019driver}  \end{tabular}                                                       & 2019          & vision-based                                                            & facial expression, eye openess, mouth open                                       & remote                    & driver drowsiness                                                                               & 25                   & 643 total events                                                                & normal, sleepy, distracted, and various other activities while driving, used in \cite{hu2021data,ed2020real}             & contains real driving conditions possibly endangering the drivers                                                                           \\ \hline
\begin{tabular}[c]{@{}l@{}}Fatigue-\\ View\footref{fn:fatigue} \cite{DBLP:journals/tits/YangYLS23}\end{tabular}        & 2022          & vision-based                                                            & head pose, eye openess, mouth closed                                        & remote                    & driver drowsiness                                                                               & -                    & 17403 yawning sets                                                              & drowsy, normal, nodding, and stretching, used in \cite{song2014eyes,drutarovsky2014eye,xie2018real,kawato2000real}                                             & office environment indoor, sitting on office chairs, simulation                                                                       \\ \hline
Drive \& Act\footref{fn:driveAct} \cite{martin2019drive}                                                & 2020          & vision-based                                                            & facial expression, head pose, eye open, mouth open                               & remote                    & autonomous driving, includes driver drowsiness        & 15                   & 12 hours of RGB, depth, IR videos, 3D body pose from 6 different views          & diverse distracting activities during both manual and automated driving conditions, used in \cite{tan2021bidirectional} & specific for autonomous driving not specific for driver drowsiness. Database is more extensive, diverse, and multi-purpose.                                                                                       \\ \hline
DMD\footref{fn:dmd}\cite{ortega2020dmd}                                                        & 2020          & vision-based                                                            & facial expression, mouth open, eyes open, head pose, yawning                     & remote                    & driver distraction, recognition includes but not only for drowsiness detection                  & 37                   & 41 hours of RGB, depth, IR videos                                               & levels of distraction, gaze allocation, hands-wheel interaction, and context data, used in \cite{abbas2022deep,DBLP:journals/tits/KotserubaT22}   & multimodal dataset for different driver scenarios. Not considering drowsiness as the only distraction factor for drivers \\ \hline
SUST-DDD\footref{fn:sust} \cite{yilmaz2022sust}                                                     & 2022          & vision-based                                                            & facial expression, eye openess, mouth open                                       & remote                    & driver drowsiness                                                                               & 19                   & 2074 RGB videos                                                                 & drowsiness and non-drowsiness                                                      & recorded under real driving scenarios, participants use their own vehicle and phones to mimic most natural and comfortable driving condition     \\ \hline
DROZY\footref{fn:drozy} \cite{massoz2016ulg}                                                      & 2016          & \begin{tabular}[c]{@{}l@{}}vision-based\\ physiological\end{tabular}    & facial expression, heartbeat variability                                         & remote, wired             & driver drowsiness                                                                               & 14                   & 3064 examples                                                                   & drowsiness and non-drowsiness, used in \cite{Reddy_Kim_Yun_Seo_Jang_2017,maior2020real}                                                        & recorded in real driving scenarios, participants use their own vehicle and phones to mimic most natural and comfortable driving condition     \\ \hline
\begin{tabular}[c]{@{}l@{}}Multi\\ channel \\ EEG\footref{fn:eeg} \cite{cao2019multi}\end{tabular} & 2019          & physiological                                                                     & EEG-based deviation in RT during lane-keeping task & wired                     & driver drowsiness                                                                               & 27                   & \begin{tabular}[c]{@{}l@{}}62 sessions derived\\ from 90 min. task\end{tabular} & deviation onset, response onset, response offset, used in \cite{cui2021subject}                                   & 23-channel EEG under simulated driving condition                                                                                                   \\ \bottomrule[0.12em]
      \end{tabularx}
    }
\end{sidewaystable*}

\section{Performance and Evaluation Metrics}
\label{sec:evaluation}
In this section, we present common evaluation metrics used to compare SOTA drowsiness detection algorithms. Popular evaluation metrics for this classification task include accuracy (Acc) \cite{deng2019real,Bakheet_Al-Hamadi_2021,Vijay_Vinayak_Nunna_Natarajan_2021}, precision \cite{garcia2012vision, tamanani2021estimation, DBLP:journals/access/KhanNKHR23}, recall \cite{garcia2012vision, tamanani2021estimation, DBLP:journals/access/KhanNKHR23}, and F1-score \cite{Bakheet_Al-Hamadi_2021,yu2017representation,tamanani2021estimation,DBLP:journals/access/KhanNKHR23} defined as follows:
\begin{equation}
    Precision=\cfrac{TP}{TP+FP},
\end{equation}
\begin{equation}
    Recall=\cfrac{TP}{TP+FN},
\end{equation}
\begin{equation}
    F1-score=2\times\cfrac{precision \times recall}{precision+recall},
\end{equation}
where TP stands for true positive accounting for the number of correct drowsiness predictions, FP stands for false positive accounting for the number of incorrect drowsiness predictions (type I error), TN stands for true negative accounting for the number of correct non-drowsiness predictions and finally, FN stands for false negative accounting the number of incorrect non-drowsiness prediction (type II errors). Detailed results in terms of accuracy and F1-score \cite{Bakheet_Al-Hamadi_2021} are calculated for assessing the detection performance. 
The mathematical formulation of accuracy is given as:
\begin{equation}
    Accuracy = \left(\cfrac{TP+TN}{TP+FN+TN+FP}\right).
    \label{eq:accuracy}
\end{equation}

In the works investigated in section \ref{sec:modern_applications}, accuracy is used as the primary performance measure. However, caution should be taken with unbalanced class distributions between drowsiness and alert labels. In cases where the evaluation data is imbalanced with respect to these labels, we advocate the use of the balanced accuracy that takes this imbalance in the class distribution into account. The mathematical equation is given by:
\begin{equation}
    Balanced-Accuracy = 0.5*\left(\cfrac{TP}{TP+FN} +  \cfrac{TN}{TN+FP}\right).
    \label{eq:balanced_accuracy}
\end{equation}

Further metrics include area under the receiver operating characteristic (ROC) Curve (AUC-ROC) \cite{papakostas2020distracted,albadawi2023real} which evaluates the classifier's ability to distinguish between drowsy and non-drowsy instances across different threshold values and Mean Squared Error (MSE) \cite{de2019detection,li2014estimation} or Root Mean Squared Error (RMSE)\cite{cui2017eeg,hachisuka2013human} which quantifies the average squared differences between predicted drowsiness levels and actual levels. Dua et al. \cite{dua2021deep} reported their results further in terms of a confusion matrix. This tabular representation comprises all four metrics (TP, TN, FP, and FN) representing the performance of the binary drowsiness detection classifier. Based on the four metrics, the authors also provide evaluation metrics such as \textit{sensitivity, specificity, precision}, and \textit{F1-score}. The mathematical formulation for \textit{specificity} is given in equation (\ref{eq:specificity}) as:
\begin{equation}
    Specificity=\left(\cfrac{TN}{TN+FP}\right).
    \label{eq:specificity}
\end{equation}
It is noteworthy, that the term \textit{sensitivity} is commonly used as a synonym to the term \textit{recall} and thus there is no need to duplicate the equation. Some works also simply reported the "Accuracy" of their algorithm. \textit{Accuracy} in this sense is dependent on the FN and FP, but unlike \textit{F1-score}, does not consider the possible imbalance in the data. Even though it is not reported in detail in most works, the accuracy can be given simply by the ratio of misclassified samples to all evaluated samples, regardless of their ground-truth label balance.

Another performance metric, when driver drowsiness detection is considered as a regression task instead of a classification task, is the case to predict the accuracy of continuous drowsiness level. Such a performance measure proposed by Wei et al. \cite{WEI2018407} is called the drowsiness index (DI). They measured the difference between the individual response time (RT) and the true alert state (alert RT) in a simulated lane-keeping task. The formulation is given by equation (\ref{eq:drowsiness_index}) as:
\begin{equation} 
    DI = max\biggl(0, \cfrac{1-e^{-\alpha(\tau-\tau_0)}}{1+e^{-\alpha(\tau-\tau_0)}}\biggr),
    \label{eq:drowsiness_index}
\end{equation}
where $\tau$ denotes the RT of the given lane-departure event, $a$ is a constant, and $\tau_{0}$ denotes the true alert RT. In their work, they continuously predict this score and related this measure to the continuous driver's drowsiness level. 

Paulo et al. \cite{Paulo_Pires_Nunes_2020} also used a leave-one-subject-out cross-validation strategy to validate their proposed classification method. Because the authors intended to perform classification without individual-dependent calibration from EEG signals, they first carried out the validation at the subject level to understand the individual contribution. By further selecting groups of subjects with major individual contributions, a better cross-subject generalizable model is created.  

These metrics in this section help to assess the effectiveness, accuracy, and reliability of drowsiness detection systems and algorithms, enabling researchers and developers to improve their models for better real-world applicability and safety. The performance metrics used are also reported in Table \ref{tab:EEG}, \ref{tab:ECG} and \ref{tab:Camera}, which show the achieved performance on bench-marking datasets for each of the three measurement techniques considered.

\section{Technical and Practical Limitations}
\label{sec:discussion}

Application areas of drowsiness detection are broad. It is of paramount importance across diverse domains due to its profound impact on safety \cite{Zhang_2017,chen2022driver,zhou2023driver}, productivity \cite{sadeghniiat2015fatigue}, and healthcare\cite{Juan-García_Plaza-Carmona_Fernández-Martínez_2021}. As extensively studied in this survey, whether on the road, in workplaces, or in critical operational environments, the ability to accurately identify and mitigate drowsiness can have far-reaching implications. In this section, we focus on identifying certain weaknesses in current algorithms and uncovering limitations in existing research. We do this with respect to two main categories, i.e., vision-based and physiological signal-based approaches. 

\subsection{Limitations on vision-based technique}

\paragraph{From the database perspective}
The increasing number of public benchmark databases for driver drowsiness detection is partially due to the rising number of fatal traffic accidents, but also following the trend towards autonomous driving \cite{martin2019drive}.  The National Highway Traffic Safety Administration (NHTSA) published the Drowsy Driving Research and Program Plan \cite{national2016nhtsa} in 2016 estimating that 2\% to 20\% of annual traffic deaths are attributable to driver drowsiness. As we have observed in previous Section \ref{sec:database}, there are several vision-based databases that can be leveraged as benchmarking systems for developing driver drowsiness detection solutions \cite{abtahi2014yawdd,weng2017driver,Ghoddoosian_2019_CVPR_Workshops,DBLP:journals/tits/YangYLS23}. However, a major limitation is that these databases often contain only simulation data or were collected under strictly controlled environments, such as indoors or in parked vehicles as reviewed in \cite{Sahayadhas_Sundaraj_Murugappan_2012}. This makes developing solutions that would work in real life scenarios more challenging. Additionally, it makes estimating the performance in real deployment questionable as the data does not represent all the variations in such scenarios. To alleviate this problem, the trend moves from simulation data to real-world data. Additional databases have been curated that have been collected under real driving conditions and incorporated various aspects of the real environment, such as different lightening and road physics \cite{chiou2019driver,martin2019drive, yilmaz2022sust}. However, it is not only tedious but also risky to collect data in a moving vehicle and it is difficult to capture all variability mimicking a real environment \cite{yilmaz2022sust}. In addition, problems such as occlusion-free capturing and accurate labeling also play an important role, which lead to works as in \cite{Ghoddoosian_2019_CVPR_Workshops,perkins2022challenges} dealing with capturing of realistic and diverse drowsiness data.

\paragraph{From the algorithmic perspective}
Most vision-based driver drowsiness detection schemes focus on observing the individual's facial attributes, such as face expression \cite{abtahi2011driver,DBLP:conf/iccp2/GhourabiGB20,vural2007drowsy,DBLP:journals/tmm/YangLMYX21,yin2009multiscale,fan2009yawning}, head position \cite{brandt2004affordable}, pupil diameter state, eye blink and eye movement (PERCLOS) \cite{DBLP:journals/sensors/ChangWCC22,garcia2012vision,wang2006vision}. With the advancement of deep learning, more accurate and efficient extraction of face and facial landmarks becomes possible, making the drowsiness detection on facial features more accurate and real-time \cite{zhao2020driver, chen2021driver,ahmed2021intelligent}. Popular face detection methods include MTCNN \cite{zhang2016joint} and RetinaFace \cite{deng2019retinaface} that allow more precise and accurate face detection under more challenging environments. Other approaches like accurate facial landmark detection in the wild \cite{zhang2014facial} also help to improve the detection of fine-grained facial expressions under varying and challenging situations. 
For integrating face recognition on devices with limited hardware resources, previous works leveraged extremely lightweight face recognition networks from knowledge distillation or model quantization \cite{DBLP:conf/icpr/BoutrosDK22,DBLP:journals/access/BoutrosSKDKK22}. To overcome the limitation of occlusion in vision-based drowsiness detection, research can benefit from methods developed for improving face detection performance under masked faces as in \cite{DBLP:journals/iet-bmt/DamerBSFKK22,DBLP:journals/pr/BoutrosDKK22}. Working on robust vision-based algorithms coping with variety of challenges faced under real life scenarios is thus a very promising future research direction. Multi angles processing \cite{9922476} and key frames selection \cite{10159554} are also the upcoming challenges for video processing in vision-based drowsiness detection. 

\paragraph{From the biased data perspective}
Proper datasets play a pivotal role in the training of deep neural networks. When datasets lack representativeness, trained models can become biased and struggle to generalize to real-world scenarios. This concern is particularly pronounced in models trained within specific cultural contexts, potentially leading to inadequate generalization due to limited racial diversity representation. This challenge is amplified in the context of driver drowsiness detection, where publicly available vision-based datasets often focus on specific ethnic groups, resulting in an incomplete picture. Ngxande et al. \cite{9042231} addressed this issue by utilizing a GAN-based method for data augmentation. They used a population bias visualization strategy to group similar facial attributes and highlight the model weaknesses in such samples. The approach involved fine-tuning the CNN model using a sampling technique for faces with subpar performance. Experimental outcomes demonstrated the effectiveness of this approach in enhancing driver drowsiness detection for ethnic groups that were underrepresented. Under the same context of data, studies \cite{worle2023induce,golz2007detection} dealing with how to induce drowsiness and collecting realistic drowsy driving data both in real traffic and under simulation are thus very important for developing robust detection algorithms.

\paragraph{From on-site hardware limitation perspective}
The requirement for a real-time and on-site driver monitoring system is crucial to avert motor vehicle accidents attributed to driver inattentiveness or drowsiness. However, onboard hardware often possesses limited computational resources. Recent advancements in deep learning, particularly model compression and distillation techniques, have made it feasible to construct compact yet highly accurate models on embedded systems \cite{jabbar2018real,Jabbar_Al-Khalifa_Kharbeche_Alhajyaseen_Jafari_Jiang_2017}, such as those integrated into vehicles. Reddy et al. \cite{Reddy_Kim_Yun_Seo_Jang_2017} proposed employing model compression to transition from a resource-intensive baseline model to a lightweight model suitable for deployment on an embedded board device. The suggested model based on facial landmarks achieved an accuracy of 89.5\% for a 3-classes classification task, operating at a speed of 14.9 frames per second (FPS) on the Jetson TK1 platform. With the European Union (EU) mandating the introduction of driver drowsiness and alertness warning (DDAW) systems for all new vehicles from 2024  \cite{worle2023induce}, the development and installation of accurate and resource-efficient algorithms to detect drowsiness in real time in the vehicle is becoming an urgent issue.

\paragraph{From data synthesis perspective}
Because of the considerable expenses associated with dataset acquisition and the lack of adequate datasets discussed earlier, we propose to use synthetic data to study the common characteristics and the various hidden impacts in data for drowsiness detection. Inspired by the insights gained from the synthetic data, we can further extend to other downstream tasks. This also allows us to uncover different causal aspects for drowsiness detection. Most works focus on finding the correlations in signal variability to the different states of drowsiness, but less on the causality aspects. We consider this to be a very promising research direction with regard to explainability. Kong et al. \cite{kong2015investigating} proposed to use Granger Causality Network to investigate driver fatigue and alertness state in EEG signals in 2015. The whole experiment included twelve young and healthy participants by recording their mental states under different simulated driving conditions and the data was analyzed by using Granger-Causality-based brain effective networks. Using such foundation models for other tasks, such as training face recognition solutions, have already gained increased interest \cite{DBLP:journals/ivc/BoutrosSFD23,Boutros_2023_ICCV}.

\subsection{Limitations on physiological signal based technique}

Physiological signal based techniques shares, to some degree, most of the limitations related to vision based techniques discussed above. Here, we focus on these limitations specifically linked to the nature of physiological signal based technique.

\paragraph{From motion artifacts perspective}
Multiple studies have investigated the correlation between EEG and driving behavior \cite{Chaabene_Bouaziz_Boudaya_Hökelmann_Ammar_Chaari_2021, wang2019spectral,Jeong_Yu_Lee_Lee_2019,cui2021subject}, consistently highlighting EEG as the most predictive physiological indicator of drowsiness. The fluctuation in band activities within EEG signals in the spectral-frequency domain offers dominant insights into varying levels of drowsiness. In contrast to vision-based detection methods that often identify drowsiness after the onset of actual sleep or during advanced drowsiness stages, technologies based on physiological signals enable the early alerting of drivers \cite{Sahayadhas_Sundaraj_Murugappan_2012}, averting potentially catastrophic accidents. However, a challenge arises from relevant signal extraction from motion artifacts by capturing biological signals either due to the motion of vehicles, individuals, or remote capture. This makes developing robust solutions what would work in real driving scenarios more challenging. Therefore, there are extensive approaches dealing with the removal of motion artifacts from EEG recordings. Relevant works include using methods like signal decomposition \cite{DBLP:conf/ieeesensors/DebbarmaB23}, wavelet decomposition \cite{DBLP:journals/sensors/HossainCRABKKAH22, 8784397}, and detrended fluctuation analysis \cite{lee2002detrended}.

\paragraph{From data transmission perspective}
Physiological signals have demonstrated their stability, reliability and precision, as they are less influenced by external factors like e.g., occlusions or variability in lightening, resulting in fewer false positive detections \cite{ji2004real,DBLP:journals/sensors/SiddiquiSBBLRD21,zilberg2022methodology}. Nevertheless, such physiological sensing methods involving cables or wired electrodes can be obtrusive and inconvenient for signal capture. Consequently, there is a noticeable trend towards wireless sensing and communication to alleviate these issues \cite{Sherbakova_Osipova_2015, Ke_Zulman_Wu_Huang_Thiagarajan_2016,Takalokastari_Jung_Lee_Chung_2011}. The hurdle lies in maintaining consistent connectivity and addressing weak signal strength in wireless setups. These challenges even exist under laboratory setup and become even more pronounced in uncontrolled environments, particularly when utilizing wearable platforms with a limited count of dry electrodes as stated by Gerwin Schalk\cite{gerwin_schalk}, a neuroscientist at New York State Department of Health's Wadsworth Center. Stable wireless data transmission and communication are thus a necessary requirement for developing high performance solutions. Signal imputation in case of data leakage is another solution to address the issue besides implementing fusion techniques. It is known that deep learning based models are often used to detect outliers or handle missing data both in handling image data or time-series \cite{7376641,DBLP:journals/ijon/ZhaoRLD23,ryu2020denoising}. 

\paragraph{From the multimodal perspective}
Often drowsiness detection scheme based on only one modality is not robust enough and thus a fusion of several complementary modalities or methods would lead to a better overall system performance \cite{daza2014fusion,yu2017representation,cheng2012driver,gwak2020investigation}. Samiee et al. \cite{samiee2014data} introduced data fusion using image-based features and driver-vehicle interaction as a strategy to address the issue of signal loss and boost the overall resilience of the driver drowsiness detection system. The outcome underlined the system's primary strengths, which encompass dependable and reliable detection and a capacity to withstand input signal losses effectively. Sedik et al. \cite{samiee2014data} investigated sensor fusion techniques in their research. Their approach involves integrating EEG, EOG, ECG, and EMG signals, resulting in enhanced system accuracy, faster detection time, and a robust drowsiness detection scheme. Signal processing techniques fuse FFT and Discrete Wavelet Transform that are applied for feature extraction and noise reduction. Various machine learning and deep learning classifiers are utilized for both multi-class and binary-class classifications. The proposed methodologies are validated through simulations in two scenarios addressing these classification tasks. The outcomes demonstrate that the proposed models exhibit high performance in detecting drowsiness state from medical signals, achieving detection accuracy of 90\% and 96\% for the multi-class and binary-class scenarios, respectively.
Recent works on multi-modal drowsiness detection systems also focused on the explainability of the algorithms it used as in \cite{hasan2024validation}. The interpretability of such models would strengthen confidence in the detection systems and increase their reliability, which is particularly important as the EU will make the installation of DDAW systems in new vehicles a legal requirement from 2024 \cite{worle2023induce}.

\begin{table*}[]
\caption{Benchmark drowsiness detection system with respect to features. Label annotations: x negative, o neutral, and + positive. We note that this assessment is based on current existing research and the negativity in any aspect of any category does not represent the potential of this category but is seen as a future research challenge. Therefore, this table is dynamic and can be changed with future research efforts.}
    \label{tab:benchmark}
    \centering
    \setlength{\tabcolsep}{3pt}
 \begin{tabular}{l|l|l|l|l|l|l|l|l|l|l}
\toprule[0.12em]
& \begin{tabular}[c]{@{}l@{}} \textbf{data} \\ \textbf{accessability}\end{tabular} & \textbf{unobstrusive} & \begin{tabular}[c]{@{}l@{}}\textbf{processing} \\ \textbf{complexity}\end{tabular} & \begin{tabular}[c]{@{}l@{}}\textbf{calibration} \\ \textbf{complexity}\end{tabular} & \begin{tabular}[c]{@{}l@{}}\textbf{noise} \\ \textbf{coupling}\end{tabular} & \begin{tabular}[c]{@{}l@{}}\textbf{motion} \\ \textbf{artifect}\end{tabular} &\textbf{occlusion} & \begin{tabular}[c]{@{}l@{}}\textbf{transmission} \\ \textbf{stability}\end{tabular} & \begin{tabular}[c]{@{}l@{}}\textbf{detection}\\ \textbf{accuracy}\end{tabular} & \textbf{practability} \\ \midrule[0.12em]
EEG       & o            & x            & x       & x        & +   & x & +         & +        & +   & x            \\ 
-wearable & o            & o            & o       & x        & x   & x & +         & x        & x   & o            \\ 
ECG       & x            & x            & x       & o        & +   & x & +         & +        & +   & x            \\ 
-wearable & x            & o            & o       & o        & x   & x & +         & x        & x   & o            \\ 
Vision    & +            & +            & +       & +        & +   & + & x         & +        & +   & +            \\ \bottomrule[0.12em]
\end{tabular}
\end{table*}

\begin{table*}
\caption{The diagram highlights the sensor types utilized for various drowsiness detection applications reported in previous works. This allows us to promptly pinpoint areas where specific sensor types are absent, promoting future research in those domains. This table does not limit the research in the field of drowsiness detection and is thus dynamic and can be changed with future research efforts.}
\label{tab:application}
    \centering
    \setlength{\tabcolsep}{3pt}
    \begin{tabular}{l|l|l|l|l|l|l}
\toprule[0.12em]
  \textbf{Application areas} & \textbf{Public transportation} & \textbf{Aviation transportation} & \textbf{Driver Monitoring} & \textbf{Workplace Safety} & \textbf{Healthcare} & \textbf{Smart Homes} \\ \midrule[0.12em]
EEG    & \checkmark  & \checkmark & \checkmark  & \checkmark  & \checkmark &\\ 
ECG    & \checkmark  & \checkmark   & \checkmark &  & \checkmark  &  \\ 
Vision & \checkmark  & \checkmark    & \checkmark  & \checkmark & &\checkmark\\ \bottomrule[0.12em]
\end{tabular}
\end{table*}

\section{Potential Directions for Research}
\label{sec:future_direction}

Table \ref{tab:benchmark} summarizes the limitations described in section \ref{sec:discussion} and provides a better overview of the relative strengths and weaknesses of the current sensing modalities and algorithms for drowsiness detection. It's important to emphasize that this assessment relies on investigated research works in this survey. Any negative aspect identified within a category should not be interpreted as a reflection of its potential; rather, it's viewed as a future research challenge.

Biosignals captured with EEG and EOG measurements are more precise \cite{lenis2016detection}, but less appropriate for use-cases as driver's drowsiness detection. Both vehicle and subject motion would strongly affect the performance and the accuracy of the feature extraction process of most proposed methods in the literature. Trends towards wearable sensors reduced the limitation of electrode-based applications of the EEG and EOG approaches to certain extent. However, headphone-based devices \cite{pham2020wake} including the biosignal measurements are also not practicable for real world driving scenarios. Drivers are not safe wearing headphone devices while driving. Therefore, future research directions involve the development of more robust algorithms mitigating these motion artifacts (e.g., by using de-trending, or signal decomposition techniques) or require more appropriate setups without affecting the drivers while driving under real world scenarios. This can include automatic learnable augmentation \cite{yang2023survey} techniques that would enrich the training data with more realistic variation.

Confusion may arise between the terms driver fatigue detection and driver drowsiness detection \cite{nayak2012biomedical, nayak2012biomedical}. 
Fatigue state describes the degree of fatigue, which may not necessarily involve drowsiness or subject falling asleep. Many individuals can experience fatigue while still remaining cognitive vigilant and are still capable of driving safely. Therefore, it is essential to prioritize the detection of drowsiness, as drivers in this state are unconscious while driving. To tackle this issue, heart rate variability (HRV) derived from ECG-signals are often employed as a feature for detection. However, ECG signals in relation towards drowsiness are often hard to draw. Changes in the parasympathetic and sympathetic activity of the body can be related to different biological processes beyond drowsiness detection. Several works \cite{lenis2016detection,meyer2006combining,adao2021fatigue} investigated the discriminative power of ECG-based features towards drowsiness and future trends hint towards more complex fusion methods \cite{chui2016accurate,golz2007feature} or more generalized self-learning models \cite{golz2007feature,mou2021isotropic}. 

Table \ref{tab:benchmark} indicates a clear advantage of vision-based solutions over a wide range of challenges faced by other two investigated measures, but also emphasized that such solutions are strongly affected by any application scenario where the measured subject might be occluded or under less favorable capture conditions \cite{Bakheet_Al-Hamadi_2021}. Features considered like the PERCLOS indicates the percentage of eye closures. This features strongly relates to the accuracy and efficiency of face detection and facial landmark detection. Video-based approaches of drowsiness detection often combine techniques for keyframe selections and final drowsiness state classification \cite{10159554}. Other works intend to face the challenge of variation in head poses as in \cite{9922476}. Therefore, we believe that future research directions in vision-based approaches should consider these relevant topics, such as accurate and fast keyframe selections and creating more robust algorithms dealing with the dynamic challenges posed in a real-world setting.

Finally, Table \ref{tab:application} shows the investigated measuring modalities successfully associated with different application areas. This allows researchers to have a clearer view to position their future research in the missing areas of drowsiness detection. However, this table should not limit the research in the field of drowsiness detection and is thus dynamic and can be changed with future research efforts.

\section{Conclusion}
\label{sec:conclusion}

Detecting drowsiness is of high significance in guaranteeing safety in various domains such as in workplaces requiring high concentration of employees, under real driving situations, in public transportation, and for aviation. Alert and well-rested employees also lead to enhanced productivity and better personal healthcare. This work is a pioneering work covering a wide application area of drowsiness detection with its modern applications and methods. We categorized our researched works into three measuring techniques, with multi-channel EEG signals, ECG signals, and vision-based detection schemes. We summarized and compared popular benchmarking databases and common evaluation metrics used to assess the performance of the developed drowsiness detection algorithms. 

We identified strengths and weaknesses in current algorithms and discussed the limitations of current research categorized under both physiological-based and vision-based approaches. We pinpointed challenges in accurate and real-time detection, in stable data transmission using wireless sensing technologies, and in building a bias-free system among others. We provide possible solutions like mitigating the bias by using synthetic, adversarial data \cite{9042231} or data augmentation \cite{Mohamed_Patel_Naicker_2023} techniques. Overcoming the hardware limitations requires model compression techniques to build small-scale but still highly accurate models, and leveraging the fusion of complementary modalities, methods, or sensors to lead to more robust and accurate detection resilient to noise or data loss. Finally, we believe that drowsiness detection remains an actively evolving field with abundant opportunities to explore, both when it comes to sensing technology and algorithmic development. The primary goal of this work is to provide an initial comprehensive survey of drowsiness detection within contemporary applications and methodologies.

\section*{Acknowledgment}
This research work has been funded by the German Federal Ministry of Education and Research and the Hessian Ministry of Higher Education, Research, Science and the Arts within their joint support of the National Research Center for Applied Cybersecurity ATHENE.

\bibliographystyle{ieeetr}
\bibliography{egbib}


\section{Biography Section}
\vspace{-1cm}
\begin{IEEEbiography}[{\includegraphics[width=1in,height=1.25in,clip,keepaspectratio]{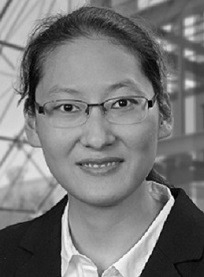}}]{Biying Fu}
received the B.Sc. degree and the M.Sc. degree in electrical engineering and information technology from University of Karlsruhe, Germany, in 2011 and 2014 respectively. She finished her Ph.D. in 2020 in informatics at Technical University of Darmstadt, Germany, on the topic "Sensor Applications for Human Activity Recognition in Smart Environments". From 2014 till today, she is working with the Fraunhofer Institute for Computer Graphics (IGD), Germany. She is currently working in the department for Smart Living \& Biometric Technologies. Starting from 2022, she also works as a half-time Professor in Informatics at Hochschule RheinMain with the major "Smart Environments". Her research interest includes intelligent human machine interaction, machine learning, and deep learning for human activity recognition with sensor data. 
\end{IEEEbiography}

\begin{IEEEbiography}[{\includegraphics[width=1in,height=1.25in,clip,keepaspectratio]{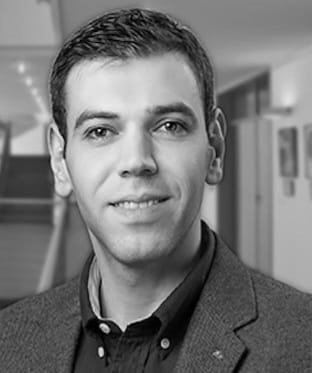}}]{Fadi Boutros} is a scientific researcher at the Fraunhofer IGD and a principal investigator at the National Research Center for Applied Cybersecurity ATHENE, Germany. Fadi received his Ph.D. in computer science from TU Darmstadt (2022) and a master's degree in "Distributed Software Systems" from TU Darmstadt (2019).  Also, he is participating in the Software Campus program, a management program of the German Federal Ministry of Education and Research (BMBF).
He authored and co-authored several conference and journal papers. His main research interests lie in the fields of biometrics, machine learning, synthetic data, and efficient deep learning. For his scientific work, he received several awards, including the CAST-Förderpreis 2019 award, the IJCB 2022 Qualcomm Audience Choice Award, and the 2022 EAB Biometrics Industry Award from the European Association for Biometrics (EAB) for his Ph.D. dissertation.
\end{IEEEbiography}

\begin{IEEEbiography}[{\includegraphics[width=1in,height=1.25in,clip,keepaspectratio]{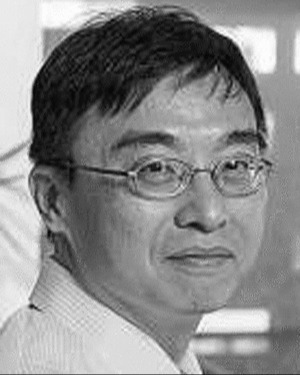}}]{Chin-Teng Lin} (Fellow, IEEE) received the B.Sc. degree from National Chiao-Tung University (NCTU), Taiwan in 1986, and holds Master’s and PhD degrees in Electrical Engineering from Purdue University, USA, received in 1989 and 1992, respectively. He is currently a distinguished professor at School of Computer Science and Director of the Human Centric AI (HAI) Centre and Co-Director of the Australian Artificial Intelligence Institute (AAII) within the Faculty of Engineering and Information Technology at the University of Technology Sydney, Australia. He is also an Honorary Chair Professor of Electrical and Computer Engineering at NCTU. For his contributions to biologically inspired information systems, Prof Lin was awarded Fellowship with the IEEE in 2005, and with the International Fuzzy Systems Association (IFSA) in 2012. He received the IEEE Fuzzy Systems Pioneer Award in 2017. He has held notable positions as editor-in-chief of IEEE Transactions on Fuzzy Systems from 2011 to 2016; seats on Board of Governors for the IEEE Circuits and Systems (CAS) Society (2005-2008), IEEE Systems, Man, Cybernetics (SMC) Society (2003-2005), IEEE Computational Intelligence Society (2008-2010); Chair of the IEEE Taipei Section (2009-2010); Chair of IEEE CIS Awards Committee (2022, 2023); Distinguished Lecturer with the IEEE CAS Society (2003-2005) and the CIS Society (2015-2017); Chair of the IEEE CIS Distinguished Lecturer Program Committee (2018-2019); Deputy Editor-in-Chief of IEEE Transactions on Circuits and Systems-II (2006-2008); Program Chair of the IEEE International Conference on Systems, Man, and Cybernetics (2005); and General Chair of the 2011 IEEE International Conference on Fuzzy Systems. He is the co-author of Neural Fuzzy Systems (Prentice-Hall) and the author of Neural Fuzzy Control Systems with Structure and Parameter Learning (World Scientific). He has published more than 450 journal papers including about 210 IEEE journal papers in the areas of neural networks, fuzzy systems, brain-computer interface, multimedia information processing, cognitive neuro-engineering, and human-machine teaming, that have been cited more than 36,000 times. His current h-index is 94, and his i10-index is 436.\end{IEEEbiography}

\begin{IEEEbiography}[{\includegraphics[width=1in,height=1.25in,clip,keepaspectratio]{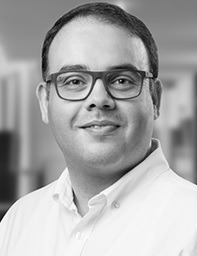}}]{Naser Damer} (Senior Member, IEEE) received the Ph.D. degree in computer science from TU Darmstadt in 2018. He is a Senior Researcher with Fraunhofer IGD. He is a Research Area Co-Coordinator and a Principal Investigator with the National Research Center for Applied Cybersecurity ATHENE, Germany. He lectures on Human and Identity-Centric Machine Learning with TU Darmstadt, Germany. His main research interests lie in the fields of biometrics and human-centric machine learning. He serves as an Associate Editor for Pattern Recognition (Elsevier), the Visual Computer (Springer), and the IEEE Transactions on Information Forensics and Security. He represents the German Institute for Standardization (DIN) in the ISO/IEC SC37 International Biometrics Standardization Committee. He is a member of the organizing teams of several conferences, workshops, and special sessions, including being the program co-chair of BIOSIG and a member of the IEEE Biometrics Council serving on its Technical Activities Committee.
\end{IEEEbiography}


\end{document}